\documentclass[10pt,twocolumn,letterpaper]{article}

\usepackage{iccv}
\usepackage{times}
\usepackage{epsfig} 
\usepackage{graphicx}
\usepackage{amsmath}
\usepackage{amssymb}

\usepackage{mathrsfs}
\usepackage{xcolor}
\usepackage{subcaption}
\usepackage{float}
\usepackage[11pt]{moresize}
\usepackage[english]{babel}
\usepackage[utf8]{inputenc}
\usepackage{csquotes}
\usepackage{makecell}
\usepackage{url}
\usepackage{balance}

\newtheorem{definition}{Definition}

\usepackage[breaklinks=true,bookmarks=false]{hyperref}

\iccvfinalcopy %

\begin{document}

\title{Monitoring Social-distance in Wide Areas during Pandemics: a Density Map and Segmentation Approach.}

\author{
Javier A. González-Trejo\\
Center for Research in Mathematics\\ CIMAT AC, campus Zacatecas, Mexico\\
{\tt\small javier.gonzalez@cimat.mx}
\and
Diego A. Mercado-Ravell$^{*}$\\
Cátedras CONACyT, CIMAT-Zacatecas\\
{\tt\small * correspondance: diego.mercado@cimat.mx}
}
\maketitle
\begin{abstract}
    
    With the relaxation of the containment measurements around the globe, monitoring the social distancing in crowded public places is of grate importance to prevent a new massive wave of COVID-19 infections. Recent works in that matter have limited themselves by detecting social distancing in corridors up to small crowds by detecting each person individually considering the full body in the image. In this work, we propose a new framework for monitoring the social-distance using end-to-end Deep Learning, to detect crowds violating the social-distance in wide areas where important occlusions may be present. Our framework consists in the creation of a new ground truth based on the ground truth density maps and the proposal of two different solutions, a density-map-based and a segmentation-based, to detect the crowds violating the social-distance constrain.
    We assess the results of both approaches by using the generated ground truth from the PET2009 and CityStreet datasets. We show that our framework performs well at providing the zones where people are not following the social-distance even when heavily occluded or far away from one camera.

\end{abstract}

\section{Introduction}

After the outbreak of the COVID-19 pandemic, the whole world witnessed how the health system was threatened to the edge of collapse. Furthermore, due to the previous lack of a vaccine or even a proper treatment for this new virus, social distancing became the only viable strategy to contain the massive contagions wave. Nevertheless, it also came to the prize of bringing the economic activities  almost to a complete stop, hence putting the social and economical stability to a sever risk. Even up to date, regardless of the successful development of vaccines against the virus, the worldwide demand is too high, and the logistics too complicated that we would need to wait some time to see the world to come back to its normality, not to forget the always present risk of a virus mutation resistant to the available vaccines.

In that context, the unavoidable need to reactivate the economy and avoid the collapse of the society, people, companies and governments have been forced to relax the strict isolation measurements, in spite of the latent risk of a new contagions wave. Accordingly, automatic social-distance monitoring has emerged as an interesting research topic that will assist the authorities to prevent massive contagions while people slowly recover their normal lifestyle.

\begin{figure}[t]
    \centerline{\includegraphics[width=0.47\textwidth]{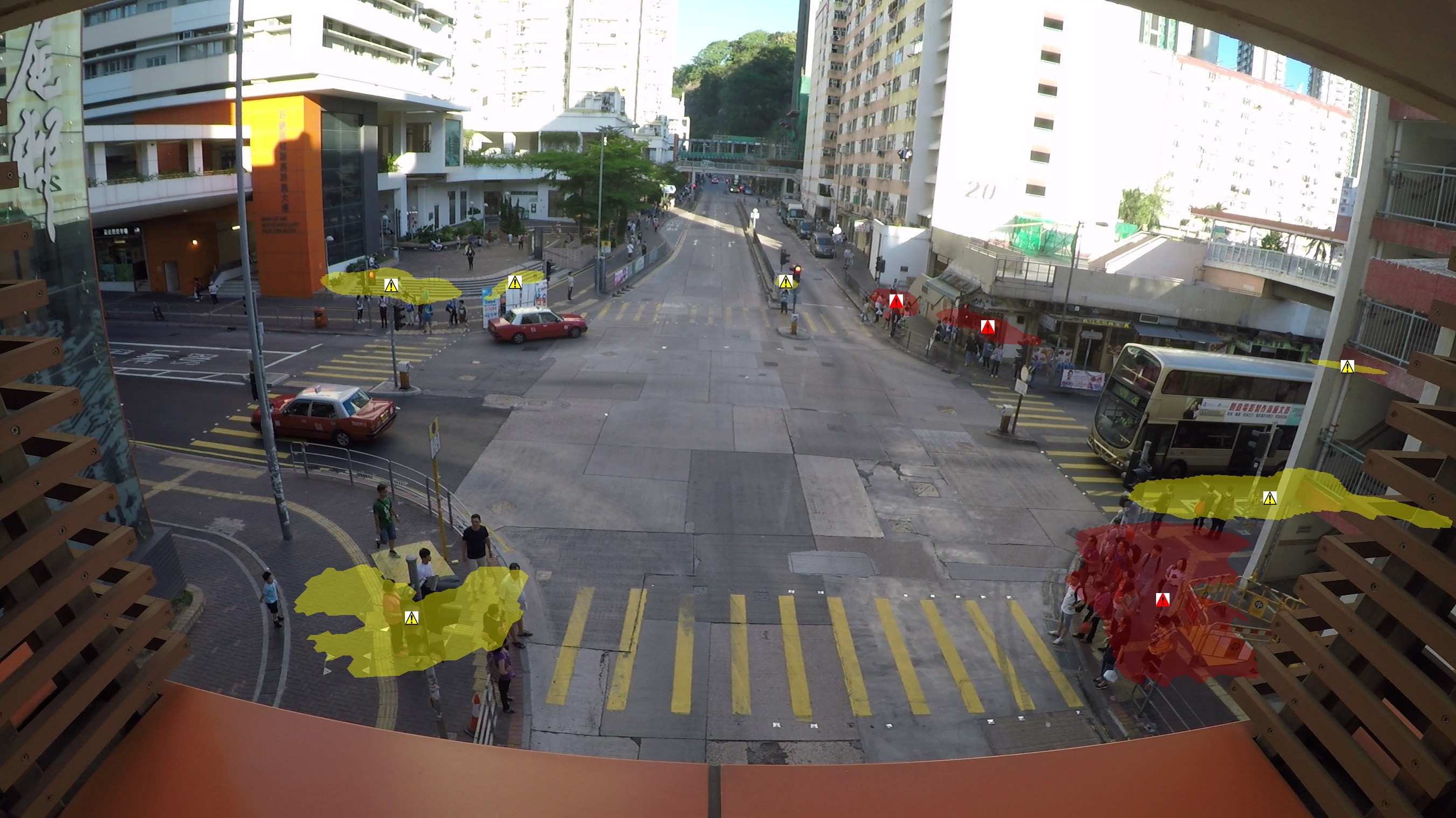}}
    \caption{Automatic monitoring social-distance in wide public areas using density maps.}
    \label{fig:no-safe-example}
\end{figure}

  Due to its actual great relevance, a few works have recently pop up in the literature in order to tackle this problem using computer vision \cite{Rezai10217514, Ahmed2021102571, punn2020monitoring, Ahamad9204934, Gupta9242628, Hou9243478}. Cristani \etal~\cite{Cristiani9138385} introduced the Visual Social Distancing (VSD) problem, as the automatic estimation of the inter-personal distance from an image, and the characterization of related people aggregations. There, the authors discuss the problem not only as a geometric one, but also considering the social implications, and even ethic aspects. Moreover, the authors identify the most common strategy for this problem, which consist in  detecting each person individually and track them along a video stream, while calculating the inter-personal distance either in image space or in ground space. Some of the classical computer vision problems involved in this kind of solution are object detection, multi-object tracking, pose estimation, homography transformations, metric scale and depth estimation, multi-view fusion, etc.
 
Up to date, all the reported works rely in the same, very intuitive principle idea, use a state-of-the-art object detector and find some sort of inter-personal distance between each individual instance. The most common detector for this task are YOLO-based (You Only Look Once) \cite{Rezai10217514, Ahmed2021102571, Hou9243478, punn2020monitoring}, but SSD (Single Shot Detector) \cite{Ahamad9204934} and Mask and Faster R-CNN (Region-based Convolutional Neural Network) have also been proposed , \cite{Gupta9242628, yang2020visionbased} respectively. Some of these works \cite{Rezai10217514, Ahmed2021102571, Gupta9242628, punn2020monitoring} further combine the detector with a tracking algorithm such as DeepSORT \cite{Wojke2017simple}, in order to improve time consistency along video streams, further improving the system precision.
 
 Following the same strategy, one of the most interesting works is the one proposed by Rezaei and Azarmi~\cite{Rezai10217514} where a new deep neural network based on YOLOv4, called DeepSOCIAL is presented for this particular task. There, the same detection and tracking framework using YOLO-based detectors and DeepSORT trackers are adopted, but the authors further asses online infection risk by statistical analyzing the spatio-temporal data from people’s moving trajectories and the rate of social distancing violations.

Although detection and tracking has proven to be a valid solution to the VSD problem, becoming the most popular, not to say the only, available kind of solution, it still presents some drawbacks inherent to the detection itself, particularly in more challenging scenarios where wider areas and larger crowds are covered and severe occlusions are present, as is common in real urban scenarios. Nevertheless, other modern deep learning techniques have proven to be more effective in such scenarios, as is the case of density maps. Density map generators are better suited for crowd counting and crowd location since they are trained to localize human head features, which are the most visible parts of a person from upper views from security cameras or drones, specially when there are severe occlusions in dense crowds or other type of visual obstacles \cite{zhang16_singl_image_crowd_count_multi}. Recently, density maps generators using Deep Learning have achieved excellent results in the detection and counting task for dense crowds, using modern techniques such as MCCN (Multi Column Neural Network) \cite{zhang16_singl_image_crowd_count_multi}. Current research on density maps not only includes the design of new architectures \cite{ranjan19_crowd_trans_networ} \cite{8804413}, but also the proposal of new loss functions specific to the task \cite{ma2019bayesian} \cite{wang20_distr_match_crowd_count}, counting from images taken from drones far above the crowd \cite{9153156}, proposing new frameworks where the data and the neural network are processed before and after the training \cite{bai2020adaptive}, and combining images taken from different type of cameras \cite{liu2020cross}.

Inspired by the recent success of density maps in the crowds' detection and counting tasks, and in contrast to the commonly used detect and track approach, we propose to tackle the VSD problem as a segmentation problem, and train Deep Neural Networks (DNN) to directly detect those groups of people not in compliance with the social-distance restriction, based only on the people's head. Also, we propose an alternative solution using density maps to detect the crowd not in compliance with the social-distance using the count information. We believe that these are unexplored and interesting alternative solutions, which may offer better performance in wide scenarios with larger crowds and important occlusions, which are common in real urban spaces. To do so, our contributions are summarized as follows:
\begin{itemize}
    \item We propose a framework to train DNN to solve the VSD problem based in either density maps and segmentation approaches.
    \item Using the head annotations in public available datasets and homography from a camera, we create the VSD ground truth by removing the social-distance conforming crowds.
    \item Based on the VSD ground truth, we propose a metric to evaluate the density map and segmentation approaches at detecting the non social-distance conforming crowds.
    \item To our knowledge, this is the first solution to the VSD problem by using density maps and segmentation, which appears as interesting alternatives for wider scenarios, where larger crowds subject to important occlusions may be present.
\end{itemize}

The article is organized as follows: in Section \ref{sec:problem_statement} we discuss the VSD problem while giving a formal definition of the social-distance, in Section \ref{sec:proposed_aprouch} we explore the framework to generate the ground truth and to train the solutions. Then, in Section \ref{sec:training_stage} we detail the training stage, while in Section \ref{sec:results_and_discussion} we present our results. Finally, in Section \ref{sec:conclusions} we give our final conclusions and future work.

\section{Problem Statement}
\label{sec:problem_statement}
\begin{figure}[t]
    \centerline{\includegraphics[width=0.47\textwidth]{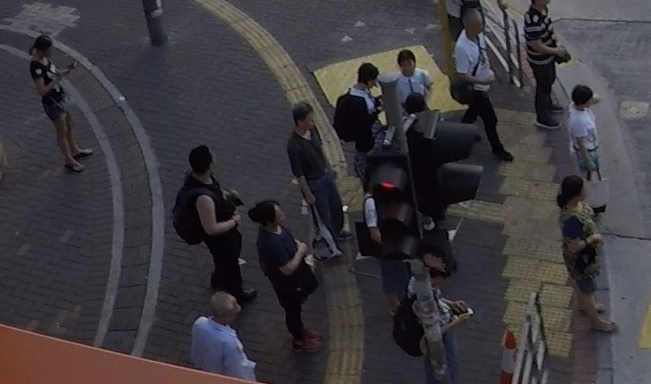}}
    \caption{Example of an urban scene where a crowd not in compliance with the social-distance is present. Only the person on the left is respecting the social-distance.}
    \label{fig:social-distance}
\end{figure}
In this paper, the objective is the automatic detection of groups of people non conforming with the social-distance in wide areas, as seen in Figure \ref{fig:social-distance}. For that matter, we consider a set of fixed cameras $\mathit{C} = \{c_1, c_2 ..., c_n\}$ each having a body reference frame $\mathbf{F}_{c_i}$ where $i \in |\mathit{C}|$. The cameras point about to the same scenario with a global reference frame $\mathbf{F}_w$, from different perspectives. The crowd appears located in $\mathbf{F}_w$, but we are only interested in the head location, since the head is the most visible part of the body given a highly occluded scenario \cite{zhang16_singl_image_crowd_count_multi}. In that regard, we make the predictions in the head's plane $\mathit{P}$ located in the frame of reference $\mathbf{F}_w$ with the center at coordinates $\mathit{P}_0 = (0, 0, h_h)$ where $h_h$ is the average height of a person. Since the images produced by the cameras operate in the image plane $\mathit{I}_i$, we need to transform the images to the global reference frame in order to know the social-distance between each person. For that, we define two transformation, $\mathbf{T}^{c_{i}}_w$ the transformation from the camera frame  $\mathbf{F}_{c_i}$ to the global frame $\mathbf{F}_w$, better known as the extrinsic camera parameters, and the transformation from the image plane $\mathit{Ii}$ to the camera frame of reference $\mathbf{F}_{c_i}$, also refereed as the intrinsic camera parameters $\mathbf{K}$. 

\begin{definition}[Social-Distance Compliance (SDC)]
\label{def1}

Let us define $H=\{\mathbf{h}_0, ..., \mathbf{h}_n \}$ as the set of the $n$ persons present in the scene, where $\mathbf{h}_i=[x_i, y_i, z_i]$ represents the $i$-th person's location in the global frame $F_w$. Then we can establish the social-distance $d_i$ for a person $i$ as the minimum inter-personal Euclidean distance $\| \cdot \|$ with respect to any other person in the scene, it is $d_i=\min_{j\neq{i}} (\| \mathbf{h}_i-\mathbf{h}_j \| ), \forall j\in [1,...,n]$. A person is considered to be in compliance with the social-distance if and only if its social-distance is bigger than a security threshold $d_t$ (normally around 2 meters), it is iif $d_i> d_t$, and is considered not in compliance (NSDC) otherwise.
\end{definition}

Then, the main goal is to develop computer vision algorithms using deep learning in order to detect those groups of people from video streams which are not in compliance with the social-distance constrain (NSDC). To do so, in the following section we describe a novel approach based on DNN segmentation.

\section{Proposed Approach}
\label{sec:proposed_aprouch}

By applying the coordinates transformations, we can project the ground truth head annotations from public available crowd datasets to the head's plane $\mathit{P}$ and remove all the persons that are in accordance with the social-distance restriction. With this new ground truth annotations, we can generate both density maps and segmentation models to train Deep Neural Network (DNN) in the head's plane $\mathit{P}$ or directly in the image plane $\mathit{I}$.

In the following, we will describe in detail the steps to generate the ground truth from the crowd counting databases and training procedures for the density map generator and the segmentation algorithm for NSDC density maps.

\subsection{Ground truth annotations}

In crowd counting, the most common form of annotation are the coordinates at the center of the visible part of the head in an image, since given a extremely dense crowd, it is the most visible part of the head \cite{zhang16_singl_image_crowd_count_multi}. By itself, this kind of annotations is not useful for detecting NSDC crowds since they do not provide the position of a person with respect to each other. Because of this, we project the annotations to the head's plane $\mathit{P}$. Thus, having the transformations $\mathbf{T}^{c_{i}}_w$ and the intrinsic camera parameters $\mathbf{K}$, the projection of the head annotation in the image plane $\mathbf{a}_\mathit{I} = (x_\mathit{I}, y_\mathit{I}, 1)$ onto the  annotation in the head's plane $\mathbf{a}_\mathit{P} = (x_\mathit{P}, y_\mathit{P}, h_h, 1)$ is given as:

\begin{equation} \label{eq:proj}
\left[
  \begin{array}{c}
        x_\mathit{P}  \\
        y_\mathit{P} \\
        h_h  \\
        1   
\end{array}\right]
   = \lambda (\mathbf{K}  \mathbf{T}^{c_{i}}_w)^{-1}
\left[
   \begin{array}{c}
        x_I  \\
        y_I \\
        1   
    \end{array}\right]
,
\end{equation}
where $\lambda$ is a scale factor. The cases where the expression $\mathbf{K}  \mathbf{T}^{c_{i}}_w$ is invertible are expressed in \cite{liu19_geomet_physic_const_drone_based}.
Since each camera in $\mathit{C}$ have a different point of view on the same scene, the redundant annotations coming from the multiple views of the same scene are used to fix the annotations position, and add extra annotations not visible by the other cameras, similar to the process described in \cite{zhang19_wide_area_crowd_count_groun}. Once we have all the annotations in the head's plane $\mathbf{a}_P$, we manually remove all the persons that are correctly following the social-distance. In other words, we remain only with the annotations $\mathbf{a}^*_\mathit{P} \subset \mathbf{a}_\mathit{P}$ which are NSDC, as stated in Definition \ref{def1}.

Now, we describe how the annotations are used to generate the ground truth density maps and ground truth segmentation for training.

\subsection{Density map generator}

Commonly, the DNN are not able to learn directly from the annotations in the head without a pre-processing \cite{zhang16_singl_image_crowd_count_multi}. For this work, we use a Gaussian kernel to blur the head annotations in both the image plane $\mathit{I}$ and the head's plane. The result is known as a density map $D_n$, which contains the location and the count of persons in the crowd in an image. The density maps cover more features of the persons head making it a more viable learning objective compared with the single point annotations. To learn how to generate these density maps $D_n$ we use the Late Fusion algorithm from \cite{zhang19_wide_area_crowd_count_groun}. It is composed by two DNN and the sampler module from the Spatial Transformers Network \cite{jaderberg2015spatial}. The first DNN is Fully Convolutional Network 7 (FCN\_7), which is used to generate density maps in the image plane. For each camera in $\mathit{C}$, a FCN\_7 is trained. Once the density maps are generated for all the cameras. The density maps are projected to the ground plane using Equation \ref{eq:proj} and the sampler module. The projected density maps count are normalized and concatenated in a single tensor to be fed into the Fusion DNN. The module learns to Fuse the projected density maps and remove the deformation caused by the projection \cite{zhang19_wide_area_crowd_count_groun}. Once the full DNN is trained, the projected density maps for that specific scenes with cameras $\mathit{C}$ can be generated. From the projected density map we obtain a mask, with the visible contours of the density map, as a visual indicator in where the NSDC crowds are located. Finally, since we have also the count information provided by the density map, we classify the crowds with a risk level and can assign "Danger" or "Warning" labels, like in Figure \ref{fig:no-safe-example}, and further purge the detection by removing the masks that have less than a threshold number of detected persons.

\subsection{Crowd segmentation}
Generate density maps involves two tasks in one, that is, while a DNN is training, it is learning how to count and where the crowd is located in the image plane $\mathit{I}$ or the head's plane $\mathit{P}$. An alternative approach to increase the detection accuracy of any DNN is to only train them to localize the crowd. In that regard, we also propose the use of segmentation to localize the crowds non conforming with the social-distance in the image plane $\mathit{I}$. We employ the ground truth density maps in the head's plane $\mathit{P}$ obtained in the previous stage, in order to generate the ground truth segmentation. As first step, we normalize the density map values and remove all the values of the density map below a threshold $t_s$, then we use the closing morphological transformation in order to create a single segmentation with no gaps between  NSDC crowds subgroups. Finally, we project the segmentation in the head's plane back to one of the cameras in the set $\mathit{C}$ using the inverse of the Eq (\ref{eq:proj}) 

The architectures considered to learn the NSDC crowds are FCN\_7 \cite{zhang19_wide_area_crowd_count_groun} and U-Net \cite{ronneberger2015u}. 
Despite the FCN\_7 not being designed for segmentation, but to produce density maps, it is still suitable for the segmentation task. FCN\_7 produces its output with only high level features hence reducing the overall visual quality of the segmentation, therefore, the ground truth segmentation have to come at the same resolution as the output of the DNN.
We use FCN\_7 as is given by Zhang and Chan ~\cite{zhang19_wide_area_crowd_count_groun}. 

Alternatively, we also tested the U-Net architecture in the crowd segmentation task. This choice comes from its decoder-encoder architecture that allows to use a ground truth segmentation that is at the same resolution as the input at training stage. This produces a better defined segmentation while, in theory, improving the precision at the task. The trade-off compared with FCN\_7 is an increase in inference and training times. 

\section{Training Stage}
\label{sec:training_stage}

In this section, we provide the technical details used for training the DNN for the task of detecting NSDC crowds and the metrics to compare the overall performance.

\subsection{Metrics}
First, we will discuss the metrics used to evaluate the methods in their respective tasks and then we discuss what to evaluate and how, for the task of detecting NSDC crowds.
For the task of crowd counting, Mean Average Error (MAE) and Mean Square Error (MSE) are the most commonly used metrics to evaluate the task \cite{kang2018beyond}. MAE and MSE are defined as follows:
\begin{equation}
    MAE =  \frac{1}{Q} \sum^Q_{q=1} |N_q - \hat{N}_q|
\end{equation}
\begin{equation}
    MSE = \sqrt{\frac{1}{Q}\sum^Q_{q=1} |N_q - \hat{N}_q|^2}
\end{equation}
where $Q$ is the total number of images in the set, $N_q$ is the ground truth count in the image $q \in Q$ and $N_q$ is the predicted total number of persons for the image $q$. MAE is used to evaluate the total count in the image while MSE highlights big errors in the count, thus MSE is usually bigger than MAE.

For the segmentation task, it is often used the \textit{Dice} score to evaluate the trained models. The \textit{Dice} score evaluates the similarity of the predicted segmentation and the ground truth segmentation by calculating the ratio of the size of the overlap between the predicted segmentation and the ground truth segmentation divided by the total area of both segmented regions. More formally, the \textit{Dice} score is defined as:
\begin{equation}
    \textit{Dice} = \frac{2 * TP }{2 * TP + FP + FN}
\end{equation}
where $TP$ are the true positives, $FP$ are the false positives and $FN$ are the false negatives, all of them measured pixel-wise in the segmentation problem.

All of this metrics are sufficient to be used to evaluate their respective task, but by themselves do not answer the question on how good are this methods at detecting the NSDC crowds while not detecting the SDC people. In this regard, we use the ground truth density maps of conforming $D_c$ and non conforming $D_n$ crowds to get how many people were correctly classified. More formally, we compute the pixel-wise $TP$, $FP$, True Negative ($TN$) and $FN$ as follows:
\begin{equation}
    TP = \hat{M} \cdot D_n
\end{equation}
\begin{equation}
    FP = \hat{M} \cdot D_c
\end{equation}
\begin{equation}
    TN = \hat{M}^{-1} \cdot D_c
\end{equation}
\begin{equation}
    FN = \hat{M}^{-1} \cdot D_n
\end{equation}
where $\hat{M}$ is the predicted segmentation region such that $\hat{M}_{i, j} \in \{1, 0\}$, and $\hat{M}^{-1}$ is the function returning the pixels not predicted as non-conforming. Having defined our TP, FP, TN and FN, we can use the traditional definitions of precision, recall, sensitivity and F1 score. Precision, recall and F1 are used to compare the methods based on how well they captured the NSDC crowds in the scene, and sensitivity for how well they do not wrongly classified the SDC crowds as NSDC.

\subsection{Datasets}
The datasets used of this paper are CityStreet \cite{zhang19_wide_area_crowd_count_groun} and PETS2009 \cite{5399556}.
\textbf{PETS2009} is a multi-view dataset designed for multiple tasks including crowd counting. In this dataset, people were told how to move and position themselves in order to challenge the solutions for the different tasks for which the dataset was designed for. In average, each frame contains 20 persons per frame. The dataset is composed of a total of 8 different views, but only three are used for the present work. Up to $794$ images extracted from the dataset are used for the purposes of this paper \cite{zhang19_wide_area_crowd_count_groun}. The resolution of each image  is $576, 768$ pixels.
\textbf{CityStreet} Is a multi-view crowd counting dataset from which $384$ annotated images are used to train the solutions here proposed. The dataset is an uncontrolled urban environment where the crowd moves at will with a total count between $50$ to $100$ persons per frame. The images have a resolution of $1520 \times 2704$ pixels, which we down sample to $480 \times 848$ for our experimentation.

\subsection{Density maps generators}
In order to train the density map generator, we need density maps $D_n$ of NSDC crowds in both the head plane $\mathit{P}$ and in the image plane $\mathit{I}$ for each camera in $\mathit{C}$. For that matter, we set the average head's position to $h_h = 1.75m$. Next, to separate the SDC head annotations from the NSDC, we used a social-distance threshold $d_t = 2m$ for two crowd datasets, the CityStreet \cite{zhang19_wide_area_crowd_count_groun} and the PETS2009 \cite{5399556}. Once we have separated the head annotations, we first produce the density maps $D^P_n$ in the head's plane $\mathit{P}$ by applying a Gaussian kernel of size $5$ and a variance $\sigma = 15$ for the CityStreet dataset, and a Gaussian kernel of size $4$ with a variance $\sigma = 15$ for the PETS2009 dataset. After that, we generate the NSDC density maps in the image plane $D^I_n$ using a Gaussian kernel of size $10$ and a variance $\sigma = 30$ for the CityStreet dataset, and a Gaussian kernel of size $4$ with a variance $\sigma = 15$ for the PETS2009 dataset. 

In the first stage, for training a FCN\_7 for each camera in $\mathit{C}$, where in this case the cardinality is set to $|\mathit{C}| = 3$ for both datasets, we set a learning rate $lr = 0.001$ using the Adam optimizer during $150$ epochs. Next, we freeze all of the FCN\_7 DNN and train only the Fusion DNN with a learning rate $lr = 1e^{-4}$ using the Adam optimizer during $150$ epochs, reducing the learning rate in case of plateau in the validation performance each $10$ epochs with a patience of $1$ and a minimum learning rate $min(lr) = 5e^{-5}$. Finally, we perform fine-tuning in the Late Fusion DNN by unfreezing the cameras FCN\_7 models with a learning rate $lr = 5e^{-5}$ using the Adam optimizer during $150$ epochs, reducing the learning rate in case of plateau in the validation performance each $10$ epochs with a patience of $0$ and a minimum learning rate $min(lr) = 5e^{-6}$. All of this hyper parameters are the same for both datasets.

At inference time, the predicted density map $\hat{D}$ is normalized. Then to generate the predicted segmentation $\hat{M}$, we saturate all the pixels values that are above a threshold equal to $\frac{20}{255}$, or we set them to $0$ otherwise. Thereafter, we select the masks that contain a count estimate bigger than $0.5$ persons and $2$ persons for the Citystreet and PETS2009 dataset respectively.

\subsection{Segmentation}
For the segmentation task, we use the $D_n$, it is the NSDC density maps in $\mathit{P}$, to create our segmentation in $\mathit{I}$. First, we normalize the density maps and set all the non $0$ pixel values to $1$. Then, we apply a morphological dilation transformation with a $7 \times 7$ ones matrix kernel $\mathbf{1}^{7 \times 7}$, and pass it trough the density map $2$ times. Next, we use the morphological erosion transformation with a kernel equal to a ones matrix $\mathbf{1}^{4 \times 4}$ for the CityStreet dataset, and $\mathbf{1}^{5 \times 5}$  for the PETS20009 dataset, also applying it trough the density map $2$ times. Finally, we project this segmentation mask back to the image plane $\mathit{I}$.

We train the FCN\_7 and U-Net models using the Adam optimizer during $150$ epochs reducing the learning rate in case of plateau in the validation performance each single epoch with a patience of $3$ and a minimum learning rate $min(lr) = 1e^{-8}$ for both datasets. As for the learning rate, we set it to $lr = 5e^{-4}$ and $lr = 0.001$ for the FCN\_7 and U-Net models respectably.

At inference time, since the segmentation per pixel is given as a value between $0$ and $1$, we saturate all the values above a threshold equal to $0.3$ for the CityStreet dataset, while for the PETS2009 dataset the best results are given by thresholds of $0.6$ and $0.9$, for the FCN\_7 and U-Net respectively.

\section{Results and Discussion}
\label{sec:results_and_discussion}
\begin{table}[b]
    \centering
    \caption{Results comparison between the different proposed methods in the CityStreet dataset}
    \begin{tabular}{|c|c|c|c|c|}
    \hline
    \textbf{Method} & \textbf{Precision} & \textbf{Recall} & \textbf{Specificity} & \textbf{F1}\\
    \hline
    \textit{Density map} & $0.889$ & $0.690$ & $0.743$ & $0.777$ \\
    \hline
    \textit{FCN\_7} & $0.882$ & $0.730$ & $0.728$ & $0.799$ \\
    \hline
    \textit{U-Net} & $0.888$ & $0.748$ & $0.728$ & $0.812$ \\
    \hline
    \end{tabular}

    \label{tab:citystreet_comparision}
\end{table}

\begin{table}[b]
    \centering
    \caption{Results comparison between the different proposed methods in the PETS2009 dataset}
    \begin{tabular}{|c|c|c|c|c|}
    \hline
    \textbf{Method} & \textbf{Precision} & \textbf{Recall} & \textbf{Specificity} & \textbf{F1}\\
    \hline
    \textit{Density map} & $0.947$ & $0.575$ & $0.614$ & $0.716$ \\
    \hline
    \textit{FCN\_7} & $0.910$ & $0.677$ & $0.4780$ & $0.776$ \\
    \hline
    \textit{U-Net} & $0.961$ & $0.761$ & $0.618$ & $0.849$ \\
    \hline
    \end{tabular}

    \label{tab:pets_comparision}
\end{table}
\begin{figure*}[ht!]
    \centering
    \begin{tabular}{cccc}
        \subcaptionbox{Ground Truth\label{fig:citystreet_a}}{\includegraphics[width = 0.23\textwidth]{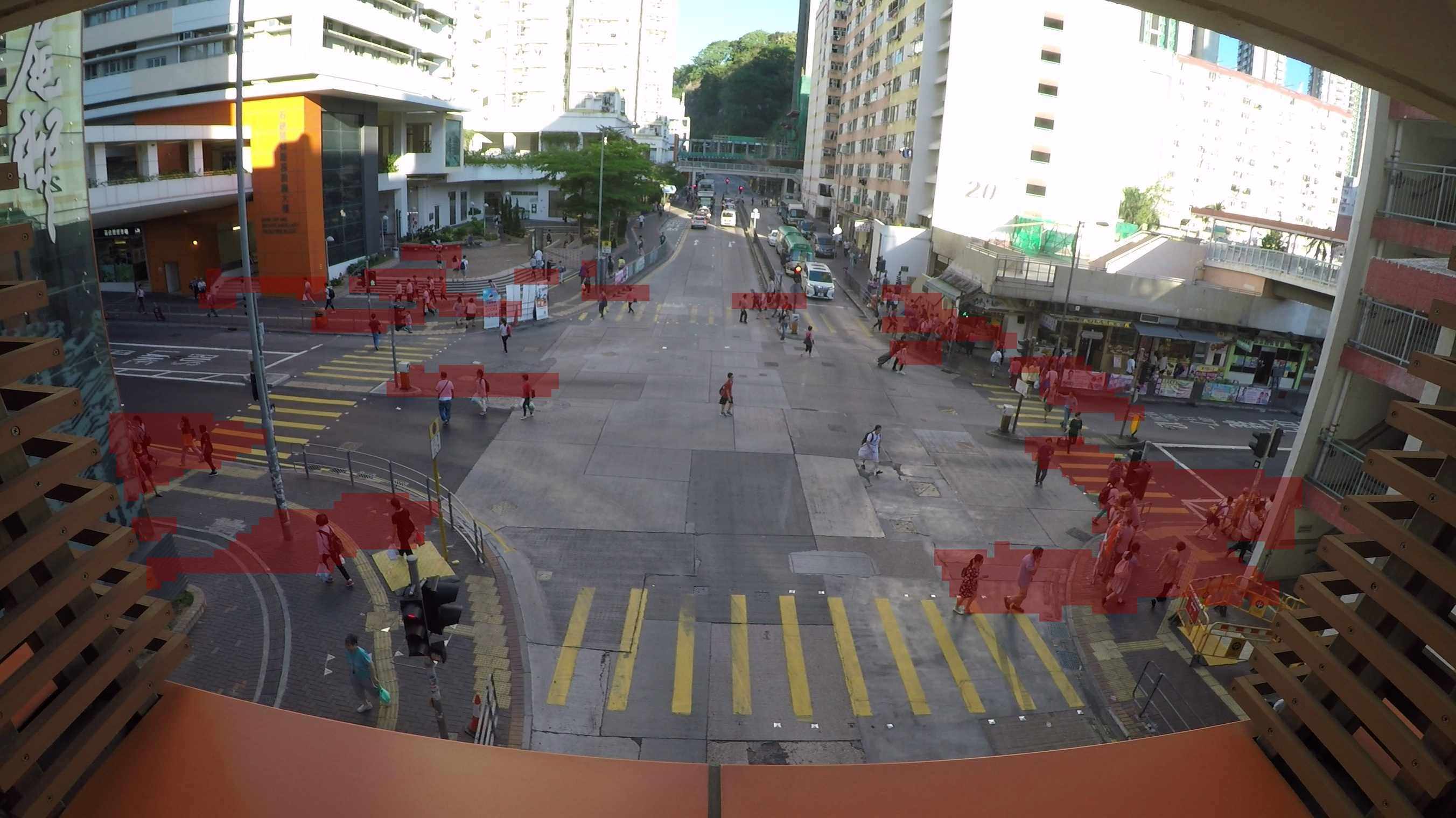}} &
        \subcaptionbox{Density map\label{fig:citystreet_b}}{\includegraphics[width = 0.23\textwidth]{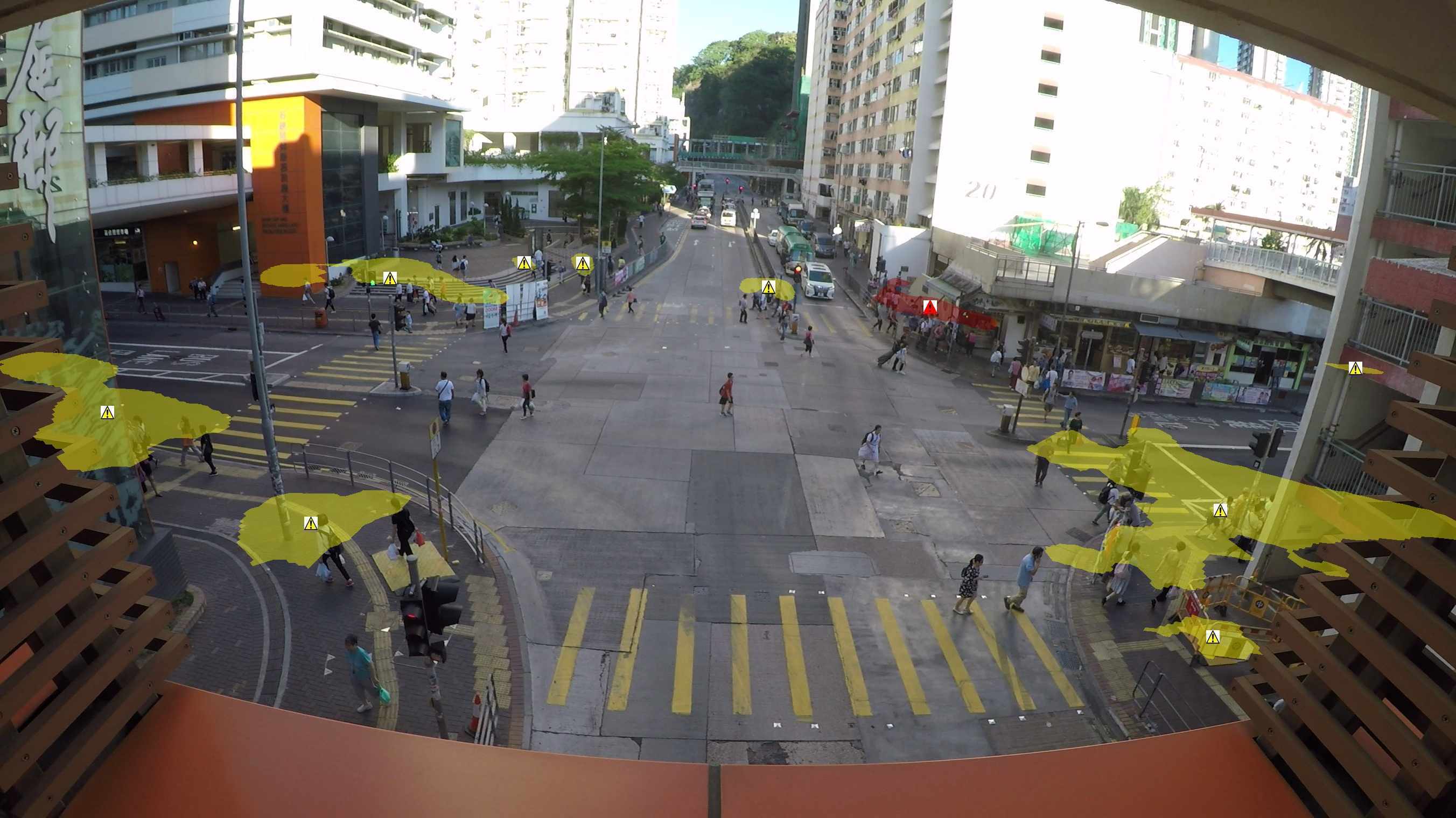}} &
        \subcaptionbox{FCN\_7 Segmentation\label{fig:citystreet_c}}{\includegraphics[width = 0.23\textwidth]{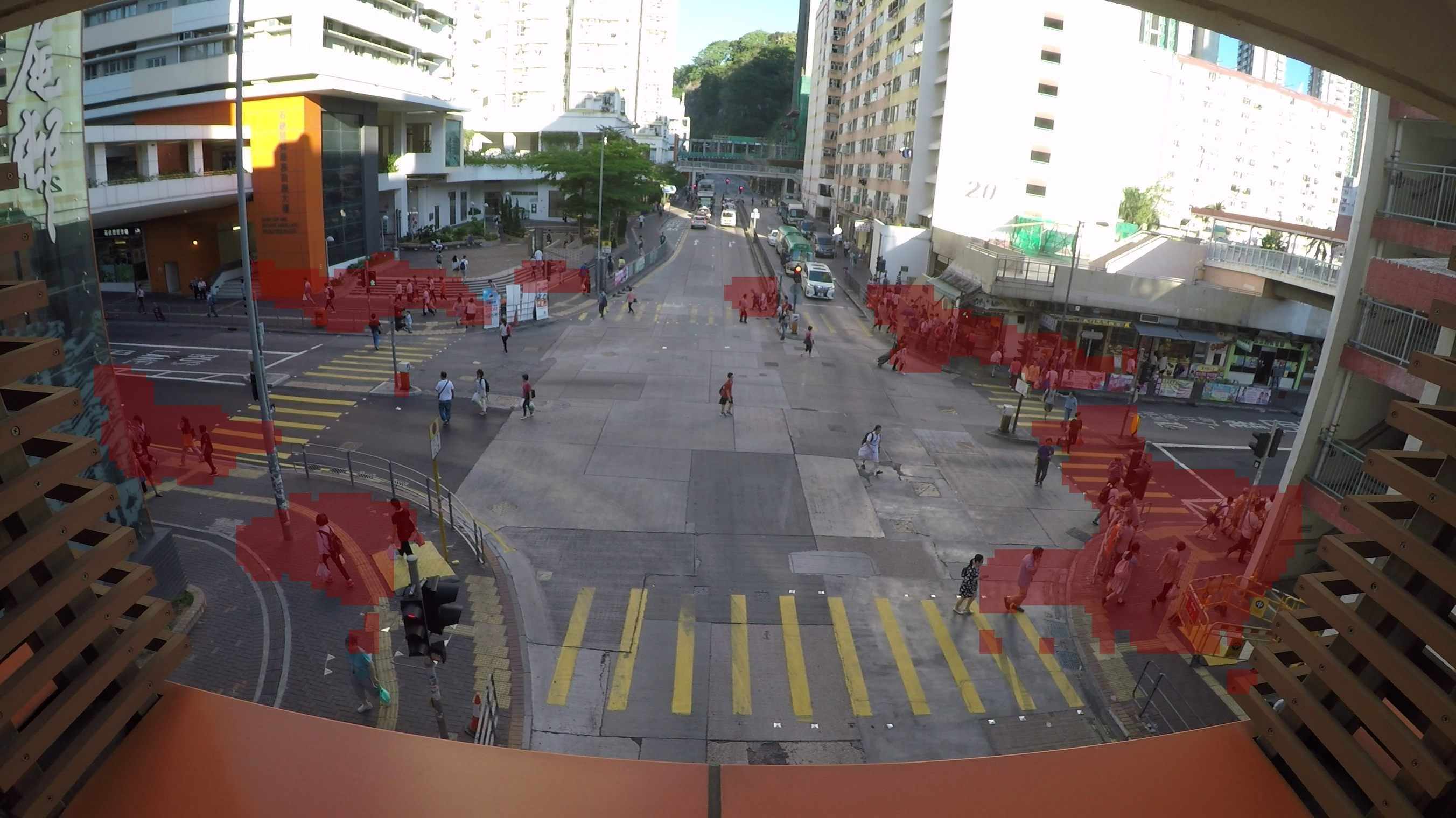}} &
        \subcaptionbox{U-Net Segmentation\label{fig:citystreet_d}}{\includegraphics[width = 0.23\textwidth]{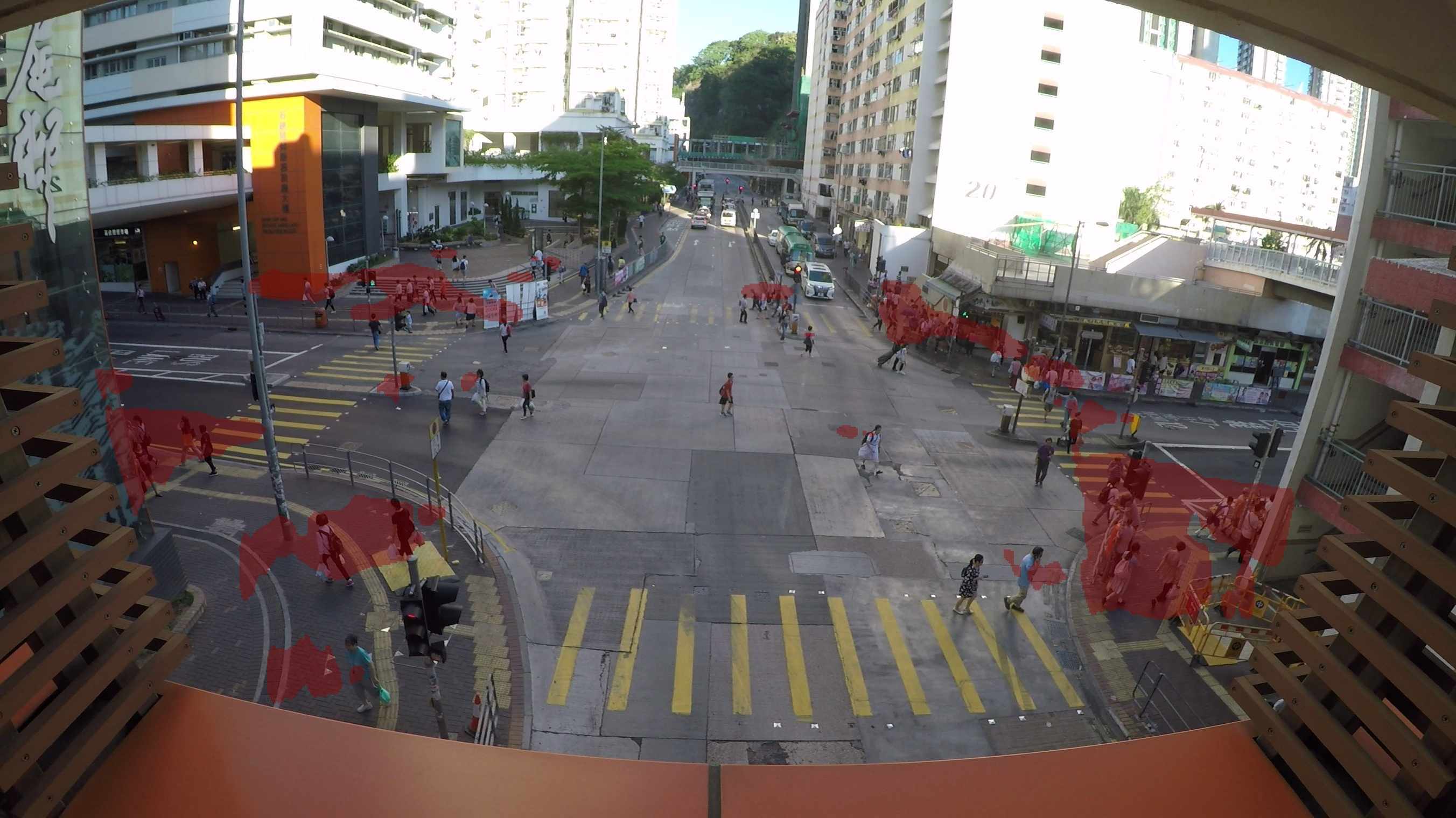}}\\
        
        \subcaptionbox{Ground Truth\label{fig:citystreet_e}}{\includegraphics[width = 0.23\textwidth]{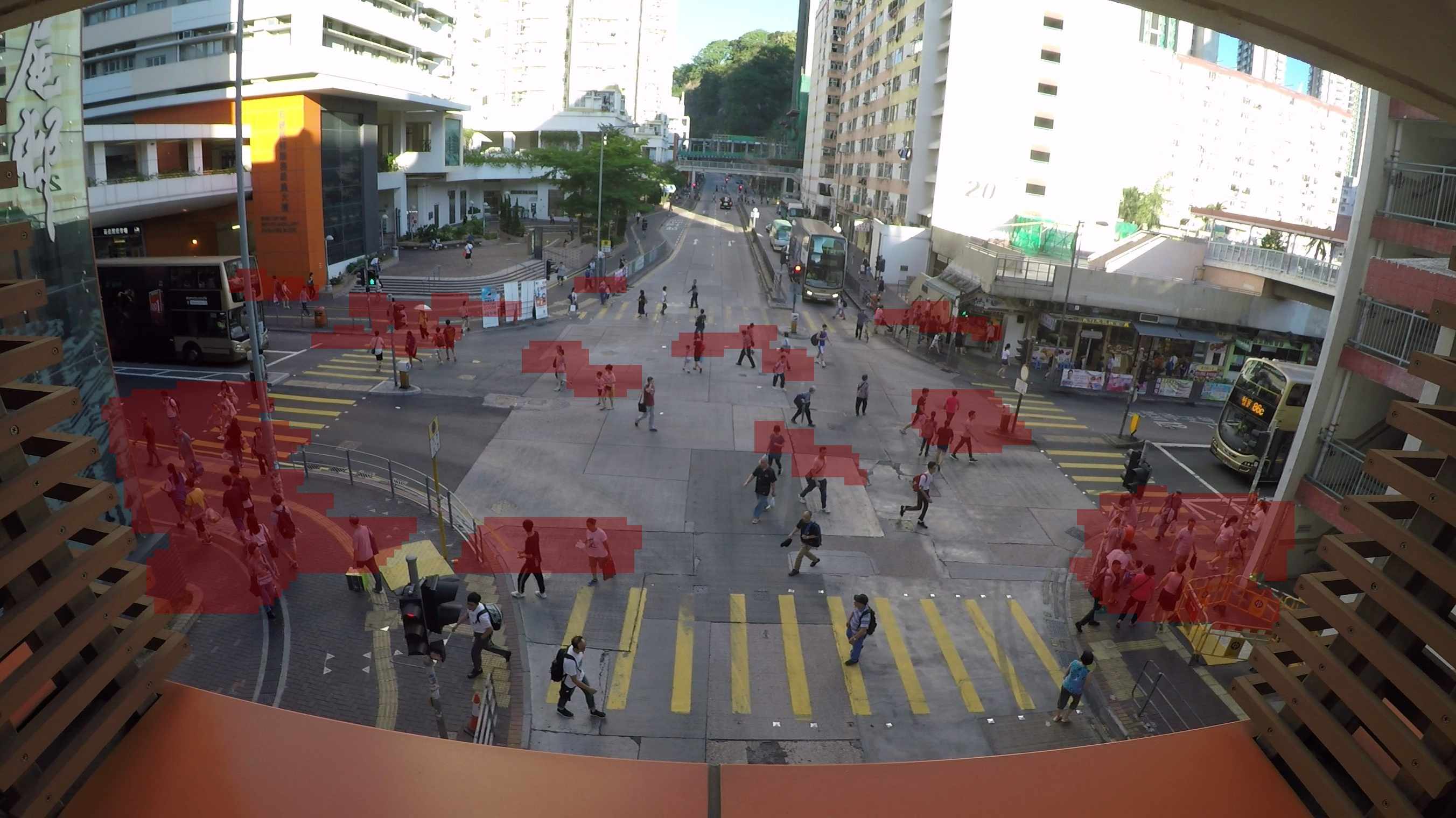}} &
        \subcaptionbox{Density map\label{fig:citystreet_f}}{\includegraphics[width = 0.23\textwidth]{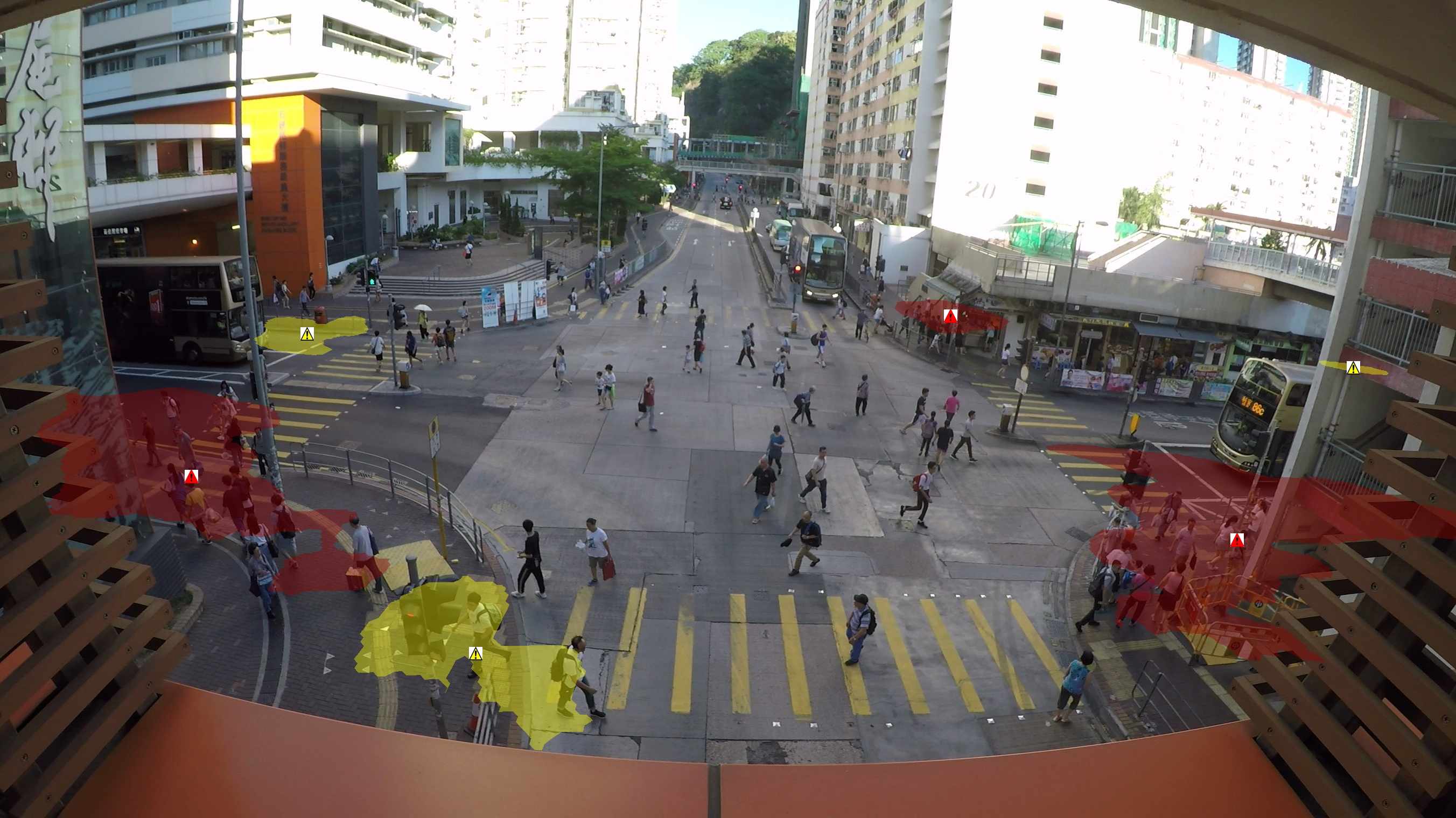}} &
        \subcaptionbox{FCN\_7 Segmentation\label{fig:citystreet_g}}{\includegraphics[width = 0.23\textwidth]{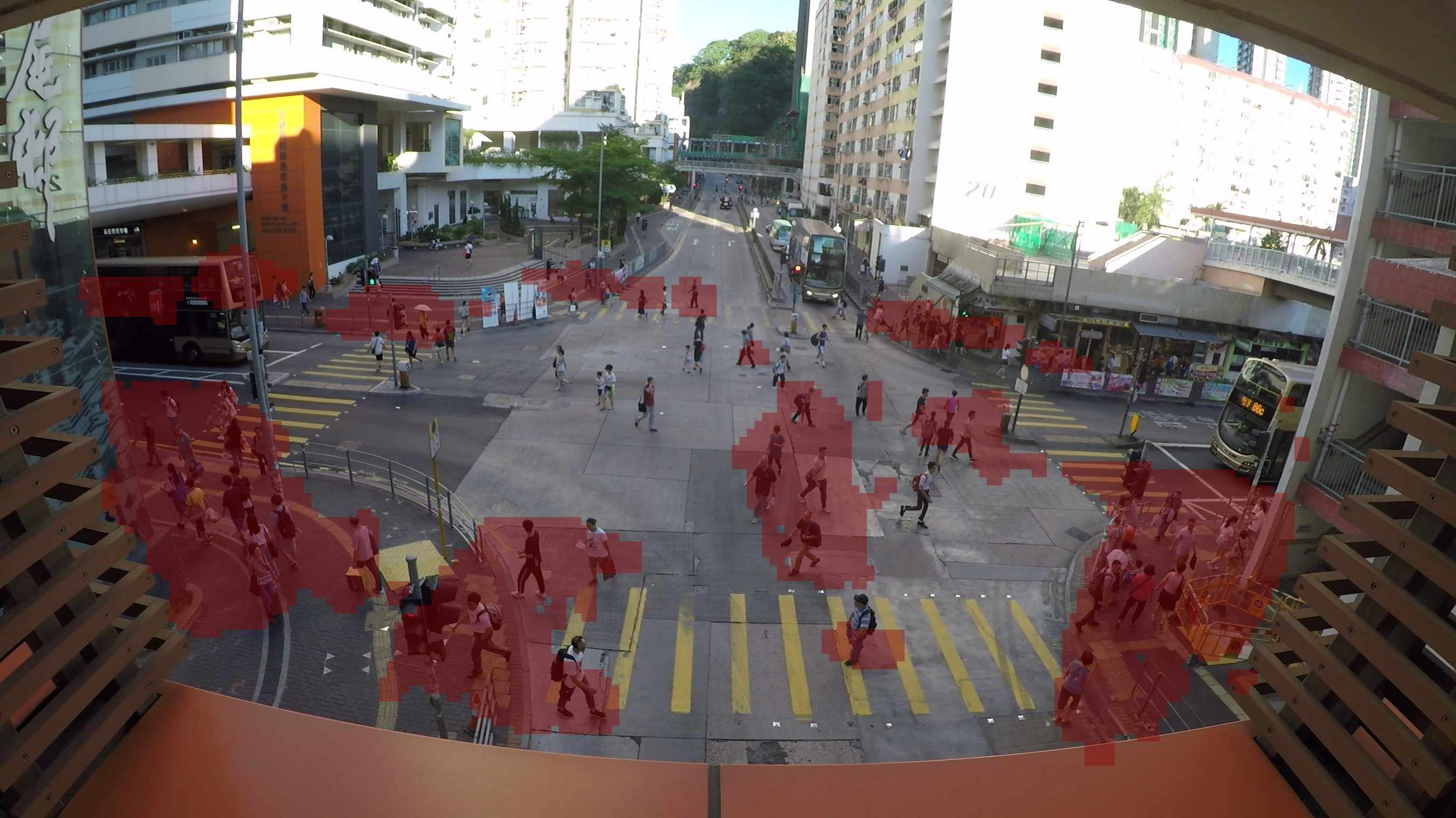}} &
        \subcaptionbox{U-Net Segmentation\label{fig:citystreet_h}}{\includegraphics[width = 0.23\textwidth]{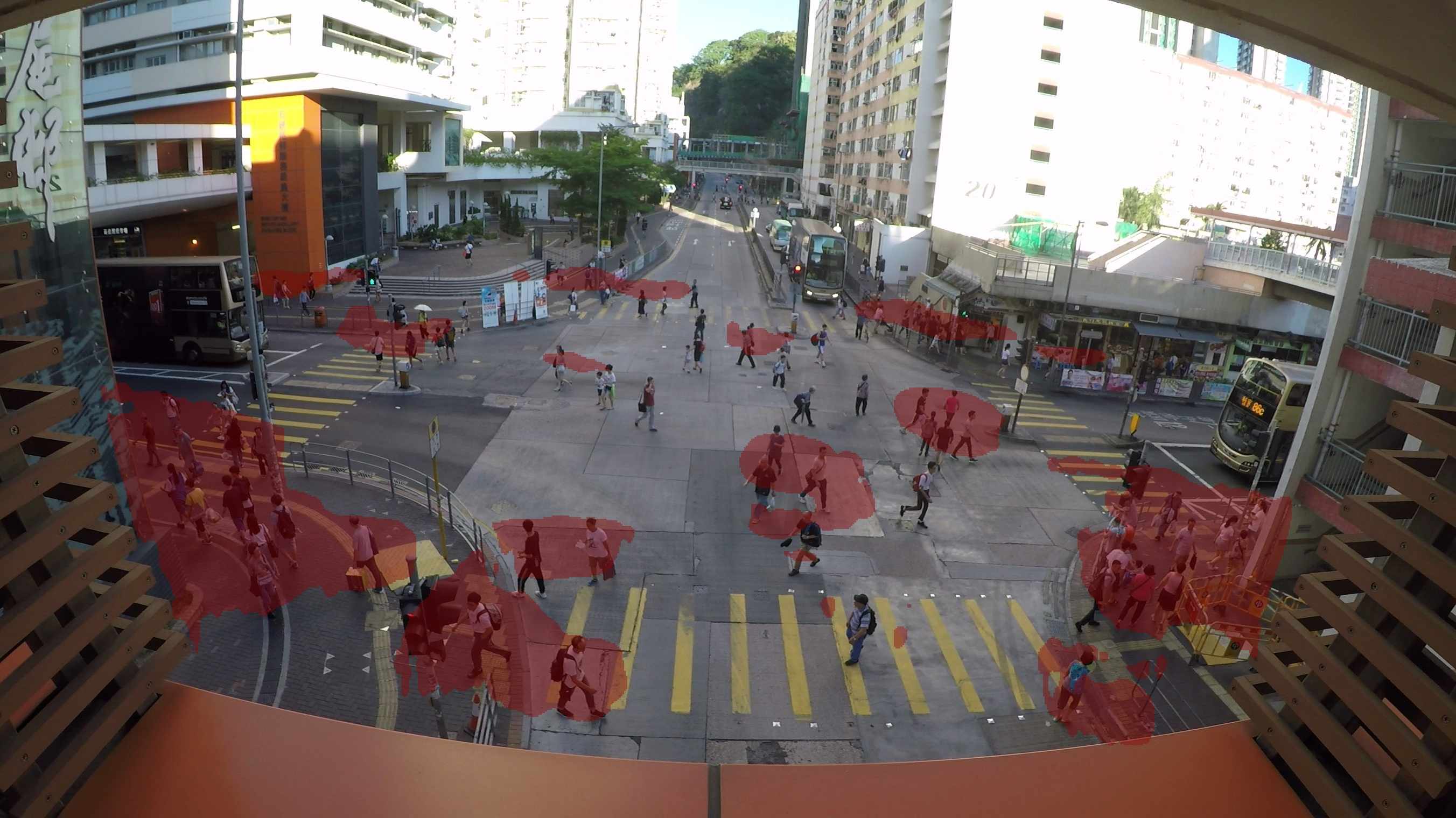}}\\

        \subcaptionbox{Ground Truth\label{fig:citystreet_i}}{\includegraphics[width = 0.23\textwidth]{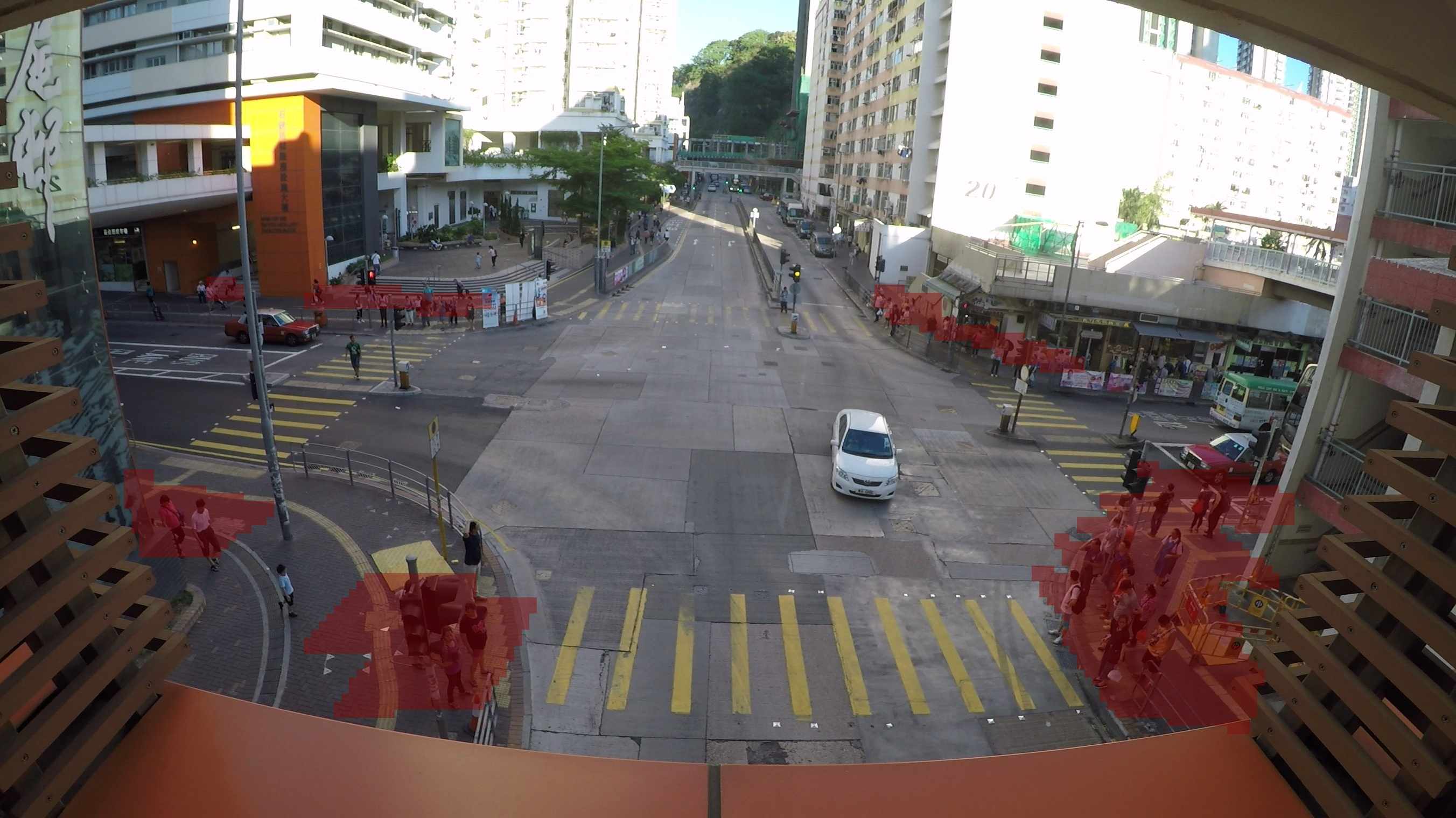}} &
        \subcaptionbox{Density map\label{fig:citystreet_j}}{\includegraphics[width = 0.23\textwidth]{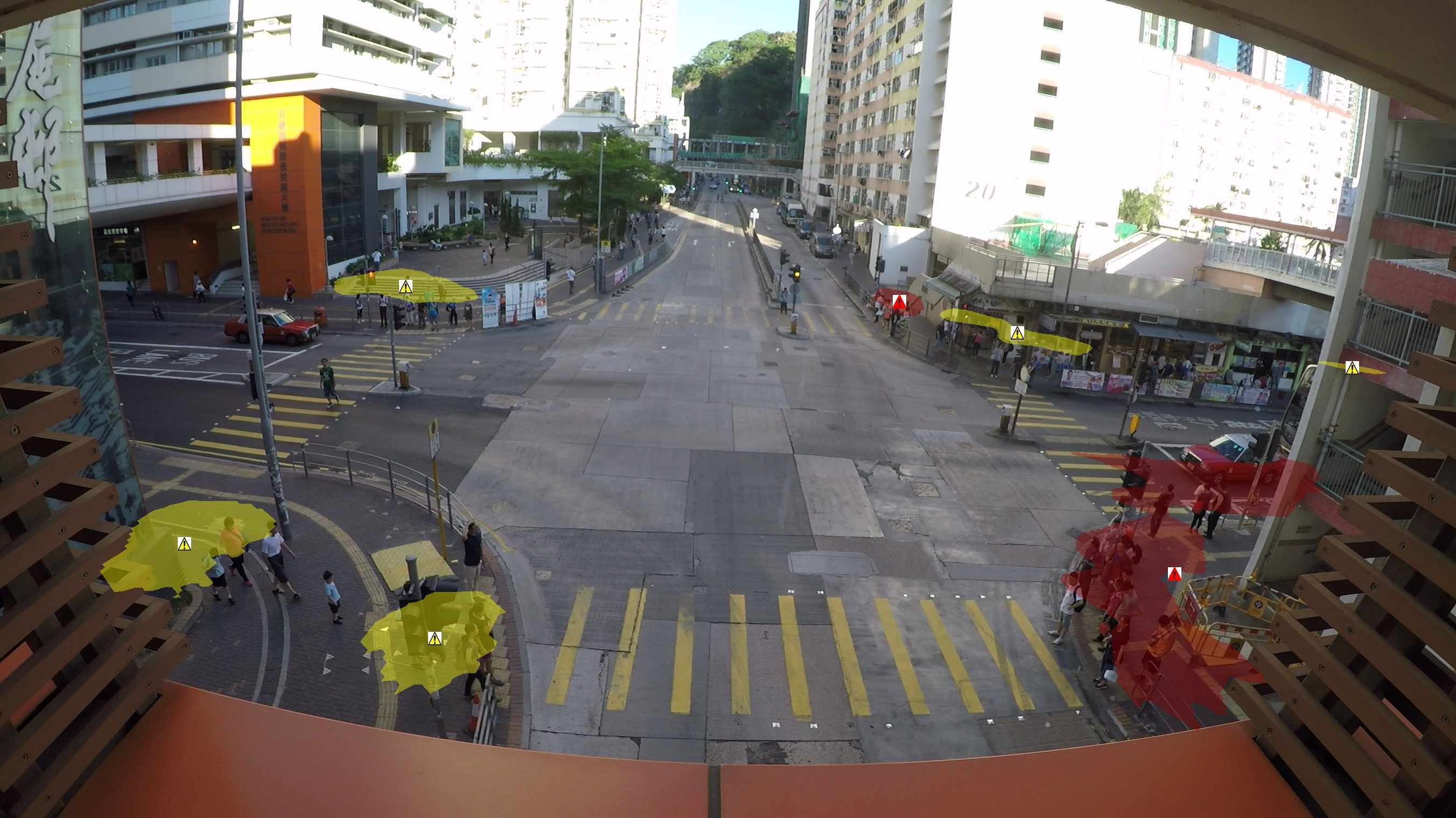}} &
        \subcaptionbox{FCN\_7 Segmentation\label{fig:citystreet_k}}{\includegraphics[width = 0.23\textwidth]{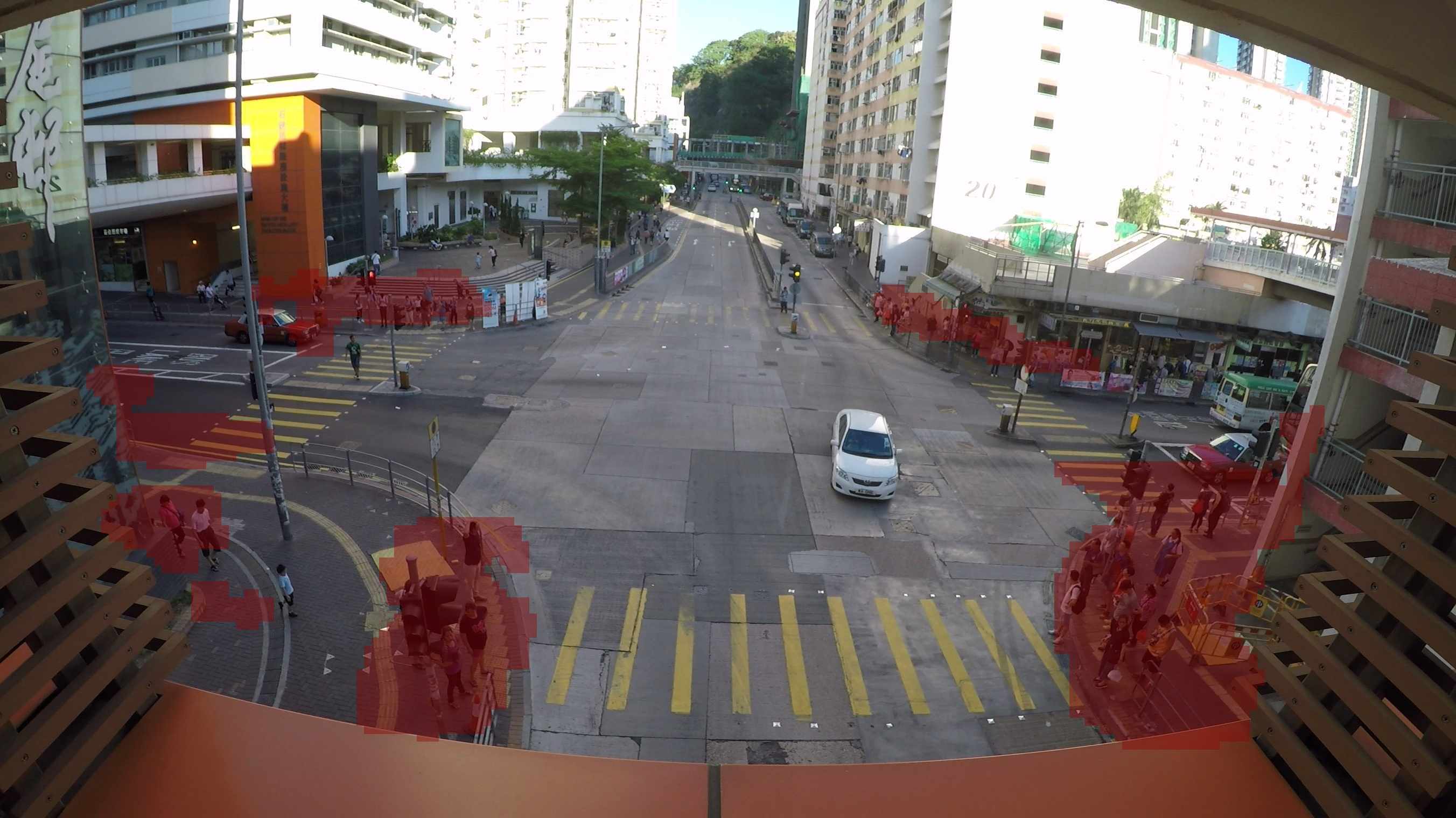}} &
        \subcaptionbox{U-Net Segmentation\label{fig:citystreet_l}}{\includegraphics[width = 0.23\textwidth]{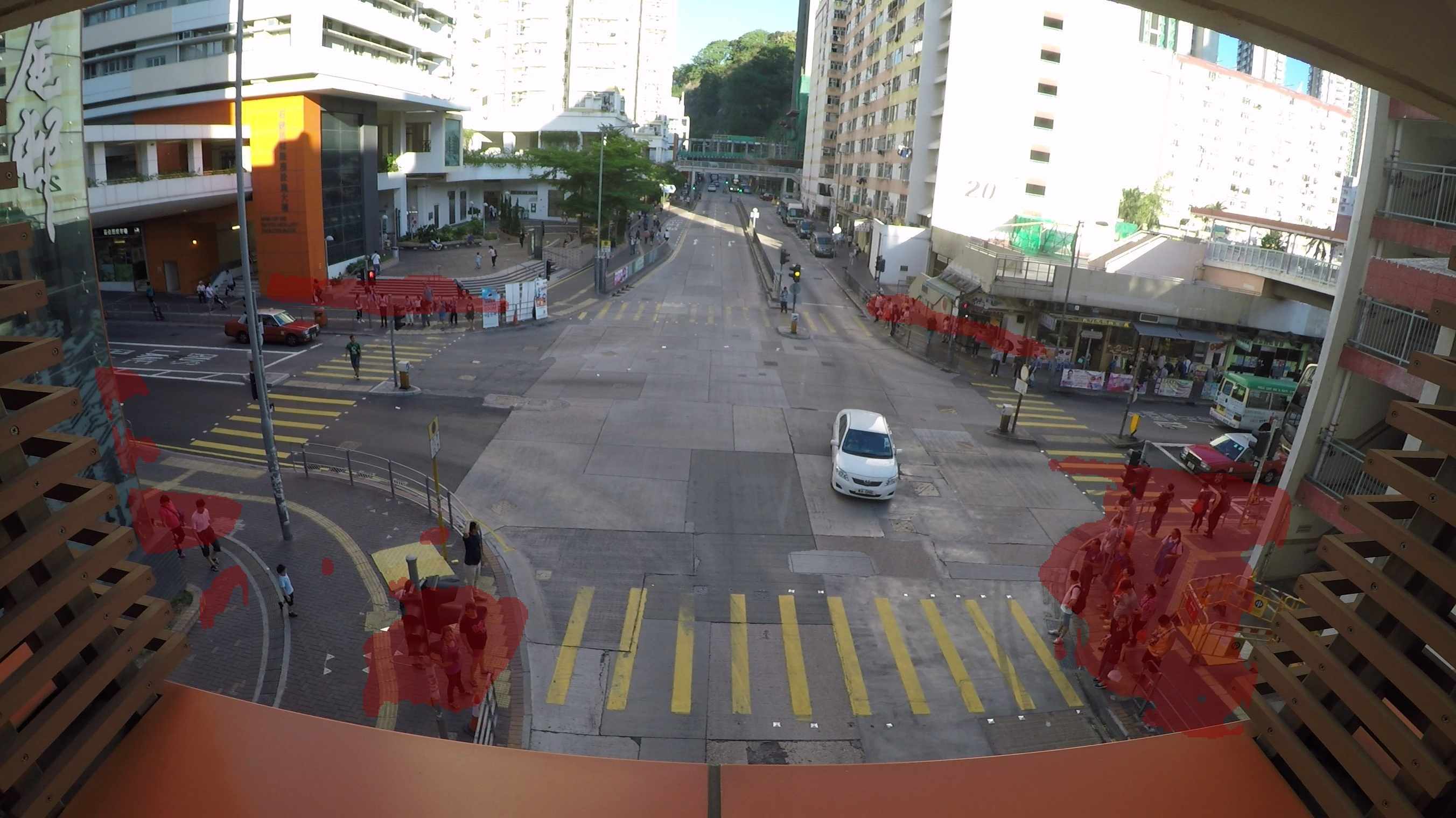}}\\

    \end{tabular}
    \caption{Results of the detection of non social-distance conforming crowds in the CityStreet dataset. We can see that the Density map based approach tends to under estimate the non conforming crowds mostly from the center. Both FCN\_7 and the U-Net perform similarly having the U-Net the edge.}
    \label{fig:citystreet}
\end{figure*}
\begin{figure*}[ht!]
    \centering
    \begin{tabular}{cccc}
        \subcaptionbox{Ground Truth\label{fig:citystreet_zoom_a}}{\includegraphics[width = 0.23\textwidth]{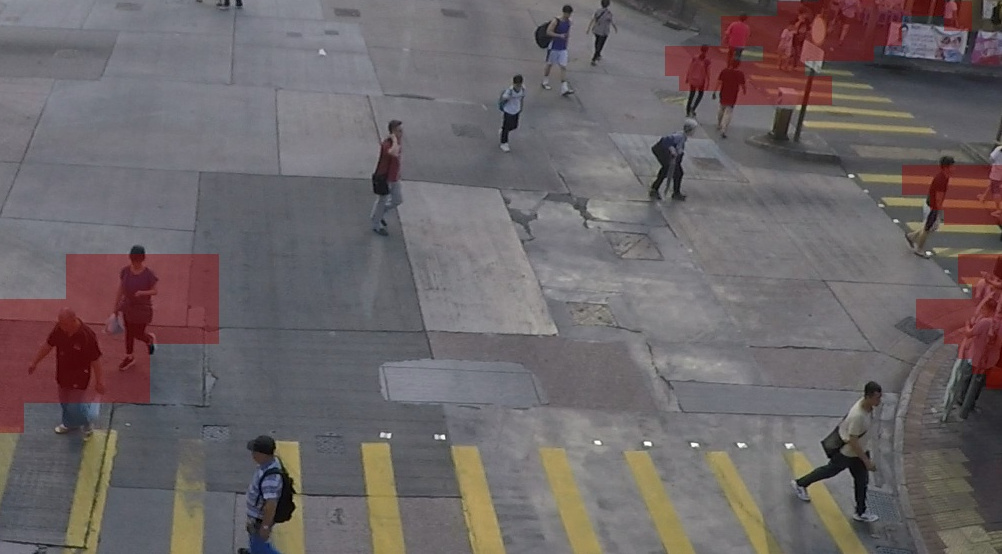}} &
        \subcaptionbox{Density map\label{fig:citystreet_zoom_b}}{\includegraphics[width = 0.23\textwidth]{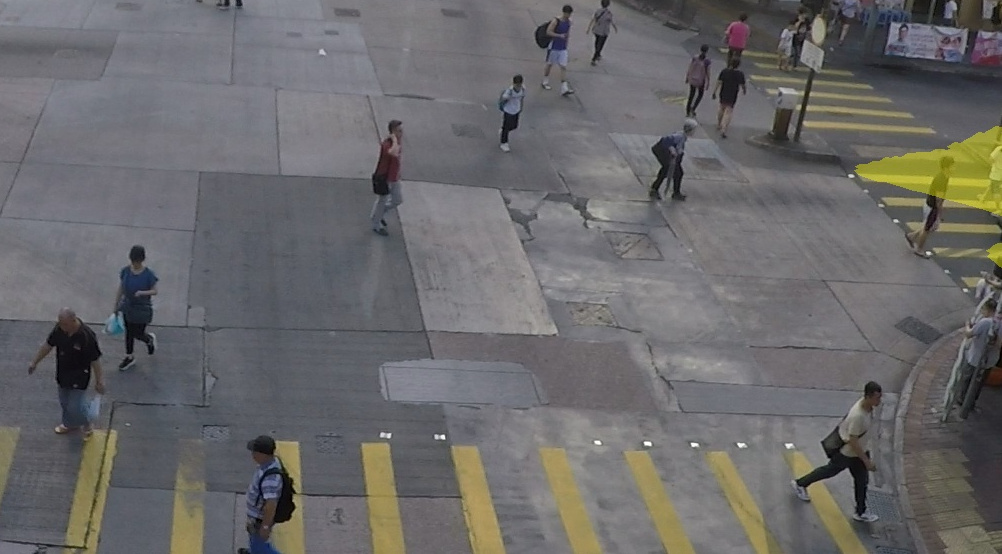}} &
        \subcaptionbox{FCN\_7 Segmentation\label{fig:citystreet_zoom_c}}{\includegraphics[width = 0.23\textwidth]{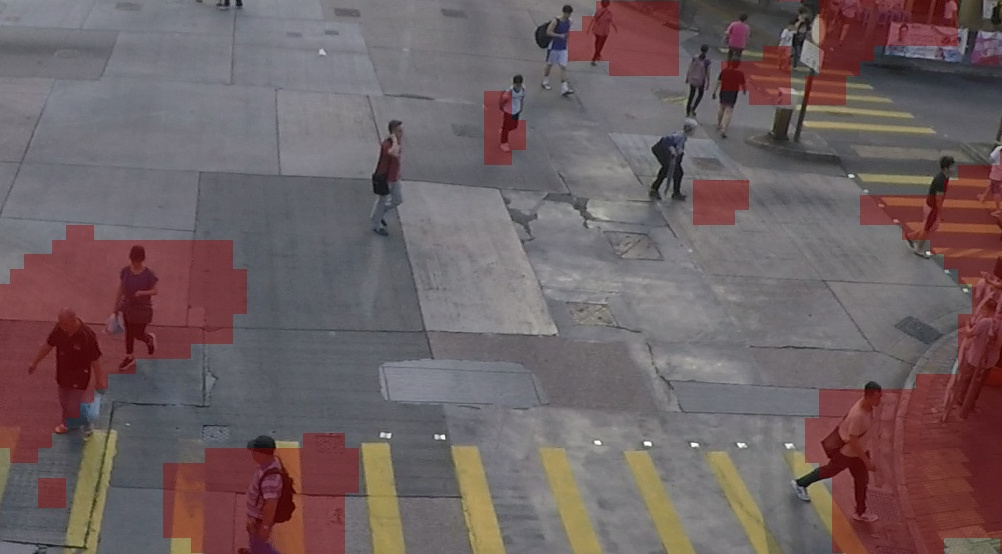}} &
        \subcaptionbox{U-Net Segmentation\label{fig:citystreet_zoom_d}}{\includegraphics[width = 0.23\textwidth]{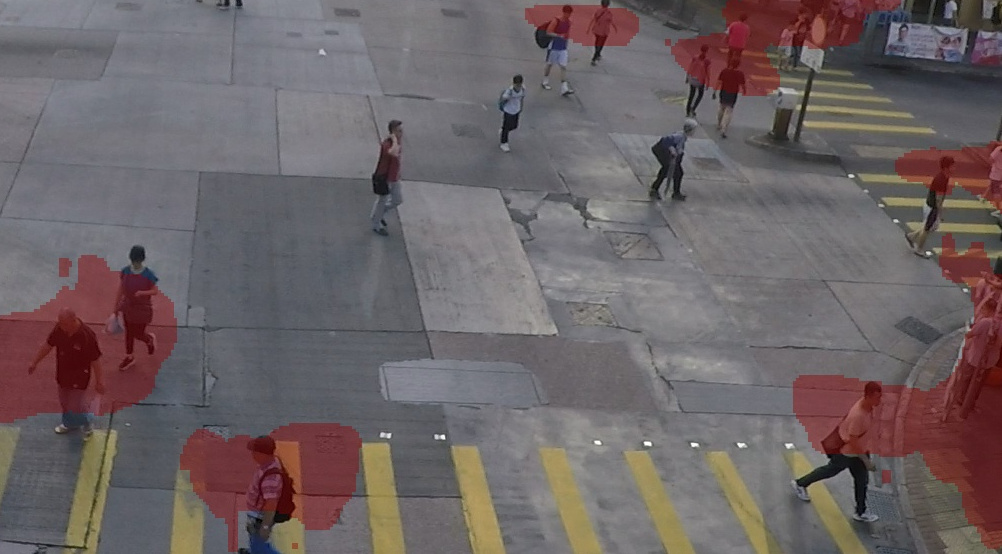}}\\
    
    \end{tabular}
    \caption{Zoomed images from the dataset CityStreet, here we can see that the U-Net performed the best out of the four approaches in this scenario despite of some False Positives.}
    \label{fig:citystreet_zoom}
\end{figure*}

\begin{figure*}[ht!]
    \centering
    \begin{tabular}{cccc}
        \subcaptionbox{Ground Truth\label{fig:pets_a}}{\includegraphics[width = 0.23\textwidth]{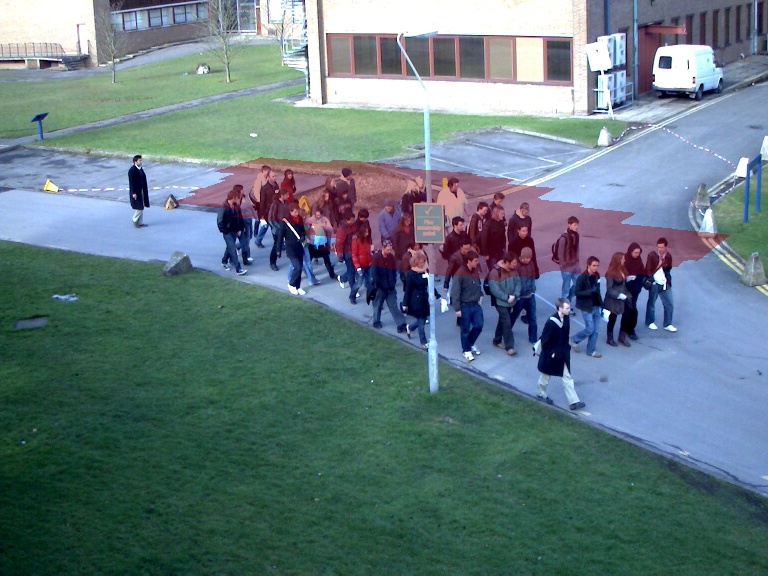}} &
        \subcaptionbox{Density map\label{fig:pets_b}}{\includegraphics[width = 0.23\textwidth]{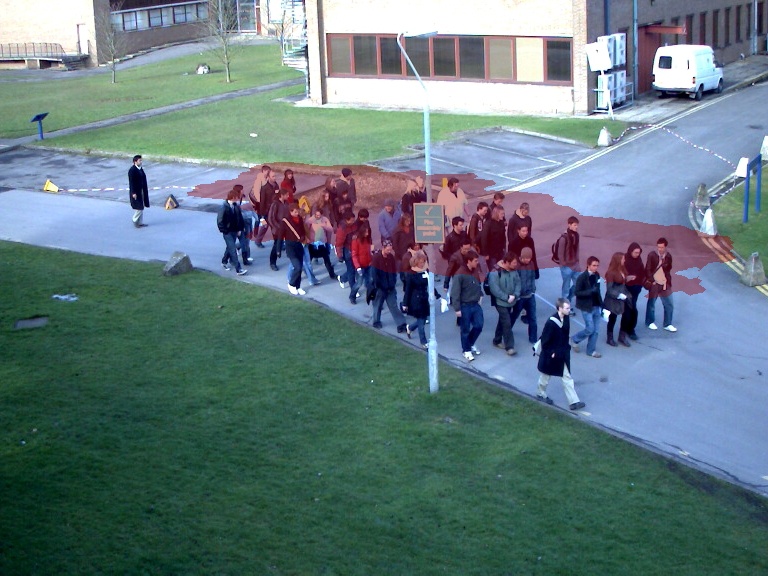}} &
        \subcaptionbox{FCN\_7 Segmentation\label{fig:pets_c}}{\includegraphics[width = 0.23\textwidth]{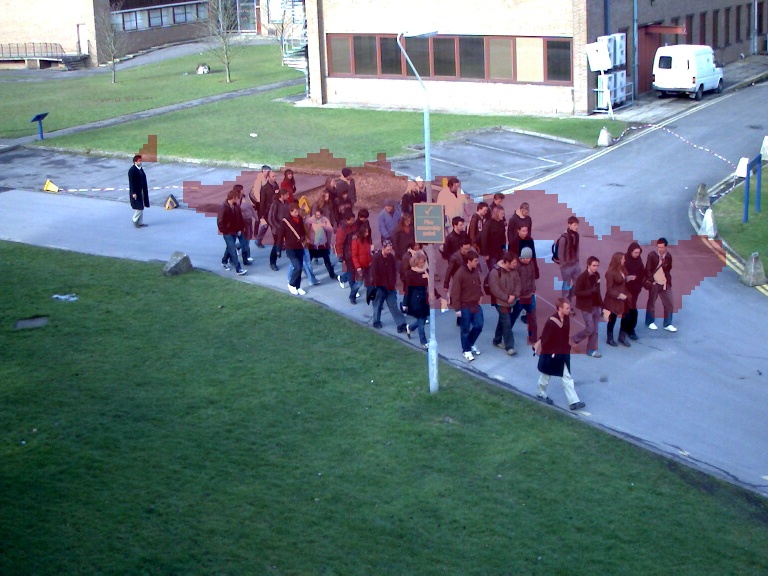}} &
        \subcaptionbox{U-Net Segmentation\label{fig:pets_d}}{\includegraphics[width = 0.23\textwidth]{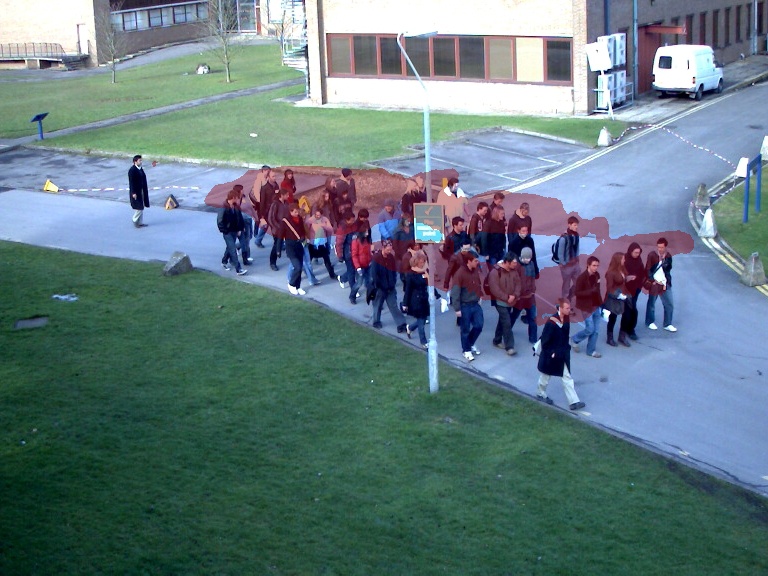}}\\
        
        \subcaptionbox{Ground Truth\label{fig:pets_e}}{\includegraphics[width = 0.23\textwidth]{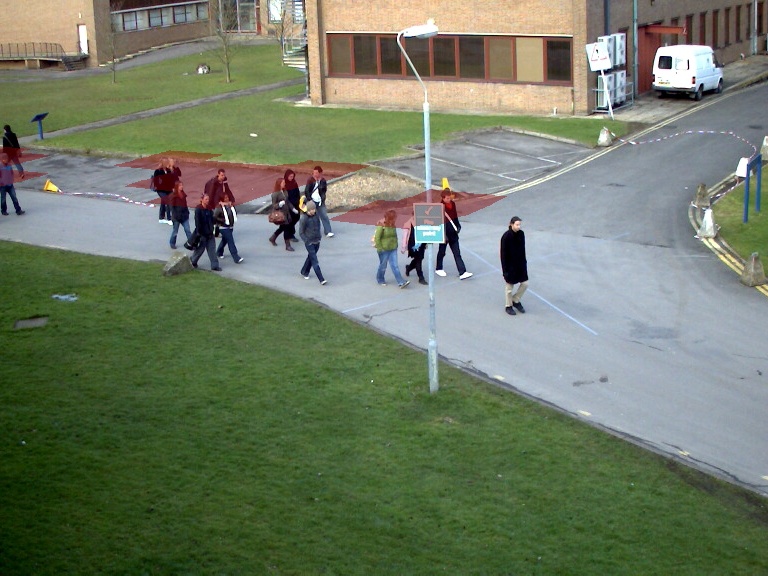}} &
        \subcaptionbox{Density map\label{fig:pets_f}}{\includegraphics[width = 0.23\textwidth]{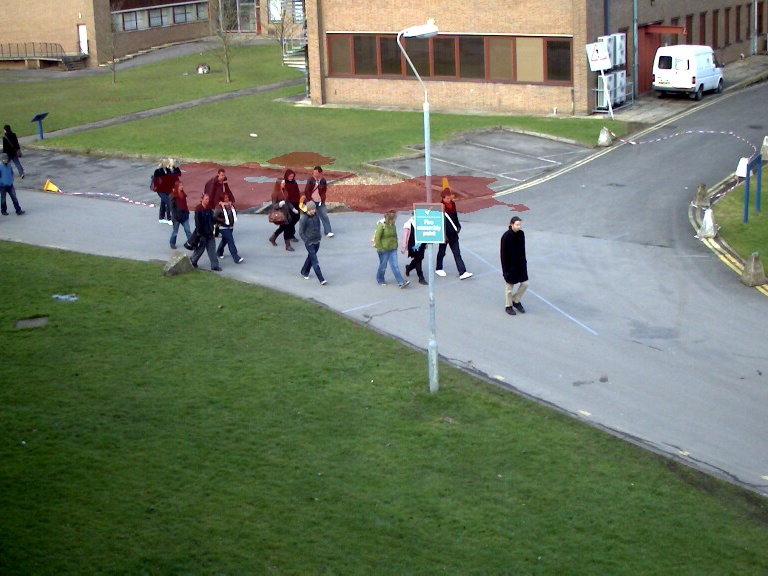}} &
        \subcaptionbox{FCN\_7 Segmentation\label{fig:pets_g}}{\includegraphics[width = 0.23\textwidth]{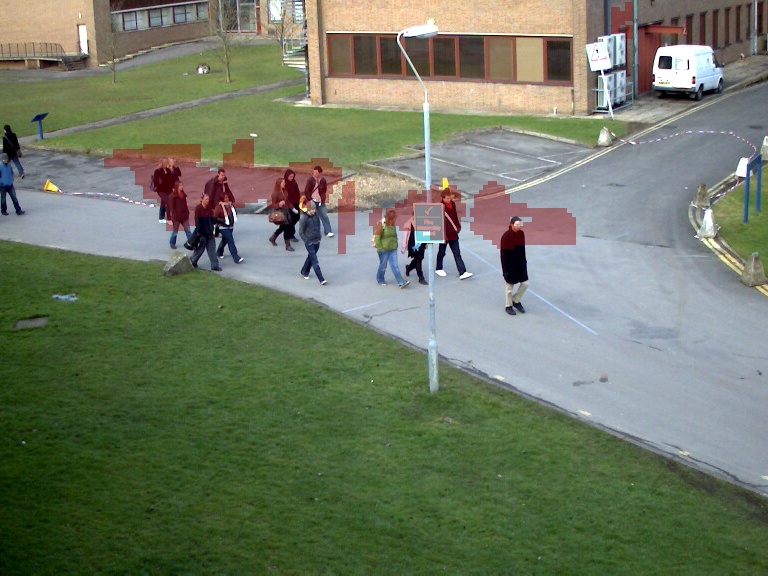}} &
        \subcaptionbox{U-Net Segmentation\label{fig:pets_h}}{\includegraphics[width = 0.23\textwidth]{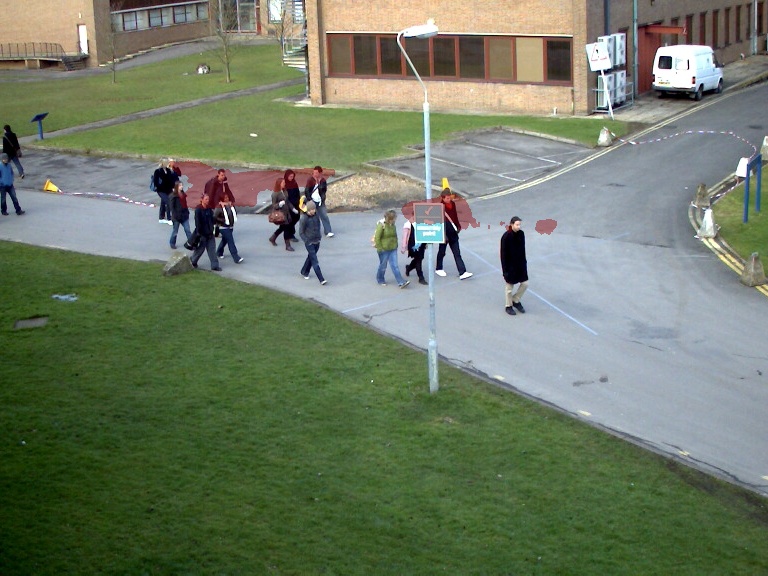}}\\

        \subcaptionbox{Ground Truth\label{fig:pets_i}}{\includegraphics[width = 0.23\textwidth]{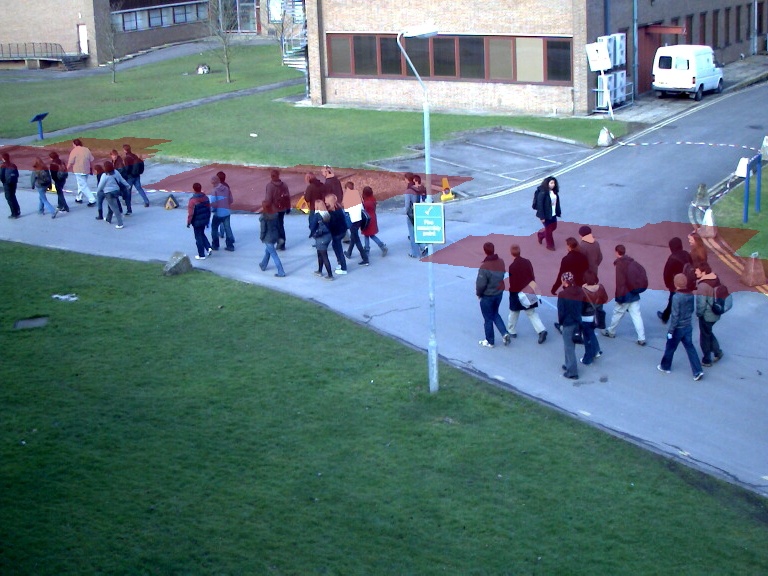}} &
        \subcaptionbox{Density map\label{fig:pets_j}}{\includegraphics[width = 0.23\textwidth]{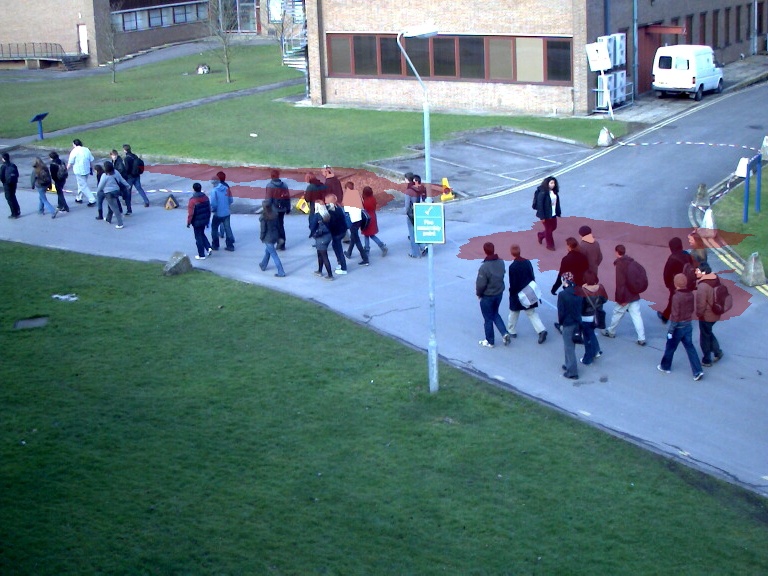}} &
        \subcaptionbox{FCN\_7 Segmentation\label{fig:pets_k}}{\includegraphics[width = 0.23\textwidth]{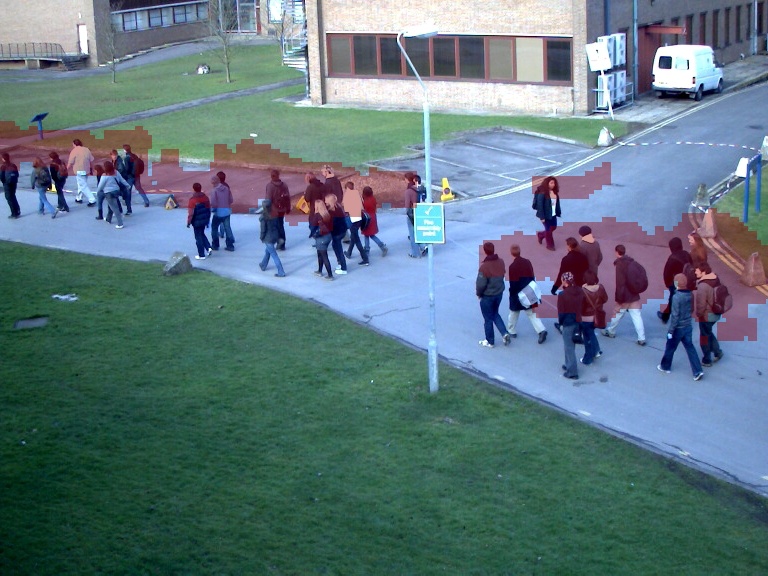}} &
        \subcaptionbox{U-Net Segmentation\label{fig:pets_l}}{\includegraphics[width = 0.23\textwidth]{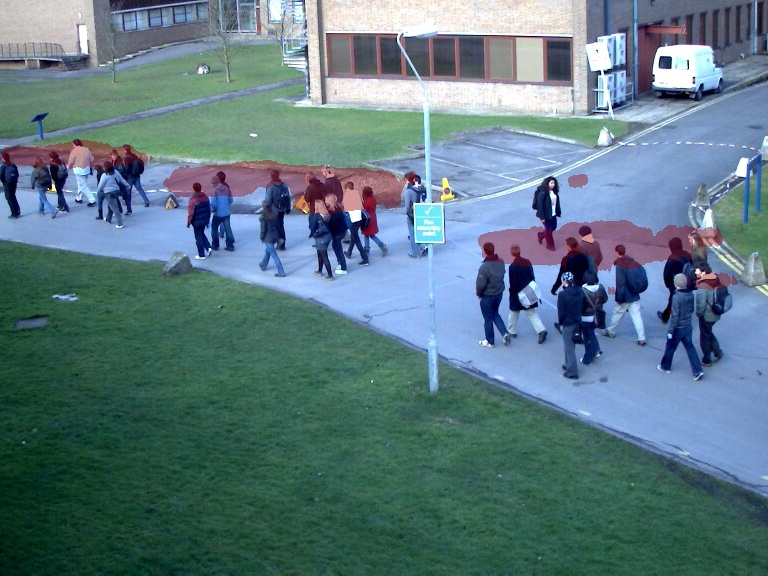}}\\

    \end{tabular}
    \caption{Results of the detection of non social-distance conforming crowds in the PETS2009 dataset. U-Net achieves the better visual results followed by the density map FCN\_7 approach.}
    \label{fig:pets}
\end{figure*}
\begin{figure*}[ht!]
    \centering
    \begin{tabular}{cccc}
        \subcaptionbox{Ground Truth\label{fig:pets_zoom_a}}{\includegraphics[width = 0.23\textwidth]{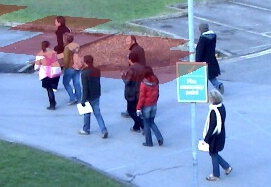}} &
        \subcaptionbox{Density map\label{fig:pets_zoom_b}}{\includegraphics[width = 0.23\textwidth]{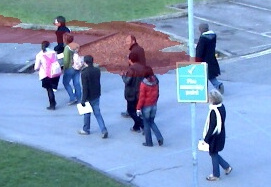}} &
        \subcaptionbox{FCN\_7 Segmentation\label{fig:pets_zoom_c}}{\includegraphics[width = 0.23\textwidth]{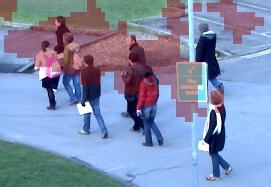}} &
        \subcaptionbox{U-Net Segmentation\label{fig:pets_zoom_d}}{\includegraphics[width = 0.23\textwidth]{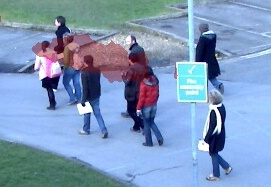}}\\
    
    \end{tabular}
    \caption{Zoomed images in the dataset PETS2009. The density map and U-Net trained are able to detect the crowds violating the social-distance while FCN\_7 trained for segmentation tends to over estimate the location of the crowd.}
    \label{fig:pets_zoom}
\end{figure*}

In Tables \ref{tab:citystreet_comparision} and \ref{tab:pets_comparision}, we present the quantitative results for all the approaches proposed in this article on the CityStreet and PETS2009 datasets, respectively.

For the CityStreet dataset, we can appreciate in Table \ref{tab:citystreet_comparision} that both the FCN\_7 and the U-Net trained for NSDC crowd segmentation performed almost the same, having the U-Net as the best overall. Although the density map approach does not stay behind the U-Net and FCN\_7 in both Precision, Specificity and F1, it is the worst at recalling all the NSDC people inside an scenario, which can be exemplified in the Figure \ref{fig:citystreet_f} where it does not detect the NSDC crowd at the center of the image. We are looking for the method that has the highest F1 score whiteout leaving the Specificity behind. 

In Figure \ref{fig:citystreet} we observe the results of three different scenarios from the CityStreet dataset, using the 3 presented methods. The density map colors with yellow and red the regions where crowds violating the social-distance constrain are found, depending on the level of risk assigned, yellow for "\textit{warning}" and red for "\textit{danger}". We can observe that FCN\_7 and U-Net performed almost equally as indicated in the Table \ref{tab:citystreet_comparision}, each having better performance in distinct situations. For example in Figure \ref{fig:citystreet_c} we see that the two persons at the lower right were labeled as NSDC while in Figure \ref{fig:citystreet_d} it only partially detects one person. On the other hand, in Figure \ref{fig:citystreet_k} the FCN\_7 model mistakes part of the ground at the left as a NSDC crowd, while in Figure \ref{fig:citystreet_l} this effect is mitigated.

More in detail, in Figure \ref{fig:citystreet_zoom} we can observe a zoomed image of the same scenario, from where it is clearer how the U-Net performs better at detecting the three persons at the center of the image as SDC, while making the same FP mistakes as the FCN\_7 model with the isolated persons at the bottom.

For the PETS2009 dataset, we observe from the results of Table \ref{tab:pets_comparision} U-Net performed better overall, while the density map approach yielded a better result in Specificity. This could be due to the number of examples of SDC persons is considerable lower with respect to the NSDC persons in the PETS2009 dataset, as seen in Figure \ref{fig:pets}, making the task more challenging. For example, in Figures \ref{fig:pets_c}, \ref{fig:pets_g} and \ref{fig:pets_k}, the FCN\_7 models wrongly detect at least one conforming person as a non conforming while almost all the NSDC crowds are correctly segmented as non conforming. In our density map approach, we can see that it is better at not classifying conforming persons, although, as seen in Figure \ref{fig:pets_j} it has some problems at classifying all the NSDC persons. As for the U-Net model trained for segmentation, we can encounter the best balance between correctly classifying NSDC crowds having some minor errors around the SDC persons from Figures \ref{fig:pets_h} \ref{fig:pets_i}, mainly due to being segmented as NSDC with low probability but removed by the threshold, leaving only the ones with higher probability.

Also, in Figure \ref{fig:pets_zoom} we see at more detail a zoomed frame from the PETS2009 dataset. From there, it can be seen that the density map and U-Net approaches almost correctly classified all the NSDC people, failing only with two persons, while the FCN\_7 model over estimates the segmentation and leaves artifacts around the conforming persons.

Finally, we show a video using the U-Net model trained for segmentation over various video sequences from the PETS2009 dataset, not including the ones used for training. The video can be found at: \url{https://youtu.be/TwzBMKg7h_U}.
\ref{tab:pets_comparision}

\section{Conclusions and future work}
\label{sec:conclusions}

In this work, we present a new framework to deal with the visual social distancing problem (VSD). Our framework proved to be useful at training Deep Neural Networks in the task of detecting non social-distance conforming crowds (NSDC) in wide areas, providing promising alternatives to the popular detect and count approach, specially in wider scenarios with more people, subject to important occlusions. 

Using the proposed framework, we presented two different approaches to solve the visual social distancing problem in wide scenarios, a density-map-based, and a segmentation-based approach. Furthermore, we evaluated the validity of these approaches for three different networks, a FCN\_7 density map generator, a FCN\_7 segmentation and a U-Net segmentation, proving that approaches based in density maps or segmentation are capable of learning the notion of social-distance by providing the ground truth annotation of only the non-conforming crowds. Moreover, we found that the U-Net segmentation showed the best performance out of the three strategies for both datasets, PETS2009 and CityStreet, achieving above $0.8$ in the F1 score and above $0.6$ in the Specificity score for both datasets. This is probably because it is a model better suited for the segmentation task. Meanwhile, the FCN\_7 model trained to detect the NSDC crowds using density maps performed better than FCN\_7 trained for segmentation in the PETS2009 dataset, possible due to the lack of enough examples of SDC people.

In future works, we aim at improving the results of our algorithms further evaluating other models. Also, we would like to provide more information about the distance in NSDC crowds in the loss function or directly in the model, and assign a level of risk accordingly. Finally, it would be interesting to monitor these crowds using mobile cameras. 

\textbf{\large Acknowledgments}\\
This work was supported by the Mexican National Council of Science and Technology CONACYT, and the FORDECyT project 296737 “Consorcio en Inteligencia Artificial”.

\balance
{\small
\bibliographystyle{ieee_fullname}
\bibliography{references}

\begin{thebibliography}{10}\itemsep=-1pt

\bibitem{Ahamad9204934}
A.~H. {Ahamad}, N. {Zaini}, and M.~F.~A. {Latip}.
\newblock Person detection for social distancing and safety violation alert
  based on segmented roi.
\newblock In {\em 2020 10th IEEE International Conference on Control System,
  Computing and Engineering (ICCSCE)}, pages 113--118, 2020.

\bibitem{Ahmed2021102571}
Imran Ahmed, Misbah Ahmad, Joel~J.P.C. Rodrigues, Gwanggil Jeon, and Sadia Din.
\newblock A deep learning-based social distance monitoring framework for
  covid-19.
\newblock {\em Sustainable Cities and Society}, 65:102571, 2021.

\bibitem{bai2020adaptive}
Shuai Bai, Zhiqun He, Yu Qiao, Hanzhe Hu, Wei Wu, and Junjie Yan.
\newblock Adaptive dilated network with self-correction supervision for
  counting.
\newblock In {\em Proceedings of the IEEE/CVF Conference on Computer Vision and
  Pattern Recognition}, pages 4594--4603, 2020.

\bibitem{Cristiani9138385}
M. {Cristani}, A.~D. {Bue}, V. {Murino}, F. {Setti}, and A. {Vinciarelli}.
\newblock The visual social distancing problem.
\newblock {\em IEEE Access}, 8:126876--126886, 2020.

\bibitem{5399556}
J. {Ferryman} and A. {Shahrokni}.
\newblock Pets2009: Dataset and challenge.
\newblock In {\em 2009 Twelfth IEEE International Workshop on Performance
  Evaluation of Tracking and Surveillance}, pages 1--6, 2009.

\bibitem{Gupta9242628}
S. {Gupta}, R. {Kapil}, G. {Kanahasabai}, S.~S. {Joshi}, and A.~S. {Joshi}.
\newblock Sd-measure: A social distancing detector.
\newblock In {\em 2020 12th International Conference on Computational
  Intelligence and Communication Networks (CICN)}, pages 306--311, 2020.

\bibitem{Hou9243478}
Y.~C. {Hou}, M.~Z. {Baharuddin}, S. {Yussof}, and S. {Dzulkifly}.
\newblock Social distancing detection with deep learning model.
\newblock In {\em 2020 8th International Conference on Information Technology
  and Multimedia (ICIMU)}, pages 334--338, 2020.

\bibitem{8804413}
V. {Huynh}, V. {Tran}, and C. {Huang}.
\newblock Danet: Depth-aware network for crowd counting.
\newblock In {\em 2019 IEEE International Conference on Image Processing
  (ICIP)}, pages 3001--3005, 2019.

\bibitem{jaderberg2015spatial}
Max Jaderberg, Karen Simonyan, Andrew Zisserman, and Koray Kavukcuoglu.
\newblock Spatial transformer networks, 2015.

\bibitem{kang2018beyond}
Di Kang, Zheng Ma, and Antoni~B Chan.
\newblock Beyond counting: Comparisons of density maps for crowd analysis
  tasks—counting, detection, and tracking.
\newblock {\em IEEE Transactions on Circuits and Systems for Video Technology},
  29(5):1408--1422, 2018.

\bibitem{liu2020cross}
Lingbo Liu, Jiaqi Chen, Hefeng Wu, Guanbin Li, Chenglong Li, and Liang Lin.
\newblock Cross-modal collaborative representation learning and a large-scale
  rgbt benchmark for crowd counting.
\newblock {\em arXiv preprint arXiv:2012.04529}, 2020.

\bibitem{liu19_geomet_physic_const_drone_based}
Weizhe Liu, Krzysztof Lis, Mathieu Salzmann, and Pascal Fua.
\newblock Geometric and physical constraints for drone-based head plane crowd
  density estimation.
\newblock In {\em 2019 IEEE/RSJ International Conference on Intelligent Robots
  and Systems (IROS)}, 11 2019.

\bibitem{ma2019bayesian}
Zhiheng Ma, Xing Wei, Xiaopeng Hong, and Yihong Gong.
\newblock Bayesian loss for crowd count estimation with point supervision.
\newblock In {\em Proceedings of the IEEE International Conference on Computer
  Vision}, pages 6142--6151, 2019.

\bibitem{punn2020monitoring}
Narinder~Singh Punn, Sanjay~Kumar Sonbhadra, and Sonali Agarwal.
\newblock Monitoring covid-19 social distancing with person detection and
  tracking via fine-tuned yolo v3 and deepsort techniques, 2020.

\bibitem{ranjan19_crowd_trans_networ}
Viresh Ranjan, Mubarak Shah, and Minh~Hoai Nguyen.
\newblock Crowd transformer network.
\newblock {\em CoRR}, 2019.

\bibitem{Rezai10217514}
Mahdi Rezaei and Mohsen Azarmi.
\newblock Deepsocial: Social distancing monitoring and infection risk
  assessment in covid-19 pandemic.
\newblock {\em Applied Sciences}, 10(21), 2020.

\bibitem{ronneberger2015u}
Olaf Ronneberger, Philipp Fischer, and Thomas Brox.
\newblock U-net: Convolutional networks for biomedical image segmentation.
\newblock In {\em International Conference on Medical image computing and
  computer-assisted intervention}, pages 234--241. Springer, 2015.

\bibitem{wang20_distr_match_crowd_count}
Boyu Wang, Huidong Liu, Dimitris Samaras, and Minh Hoai.
\newblock Distribution matching for crowd counting.
\newblock In {\em Advances in Neural Information Processing Systems}, 2020.

\bibitem{9153156}
Q. {Wang}, J. {Gao}, W. {Lin}, and X. {Li}.
\newblock Nwpu-crowd: A large-scale benchmark for crowd counting and
  localization.
\newblock {\em IEEE Transactions on Pattern Analysis and Machine Intelligence},
  pages 1--1, 2020.

\bibitem{Wojke2017simple}
Nicolai Wojke, Alex Bewley, and Dietrich Paulus.
\newblock Simple online and realtime tracking with a deep association metric.
\newblock In {\em 2017 IEEE International Conference on Image Processing
  (ICIP)}, pages 3645--3649. IEEE, 2017.

\bibitem{yang2020visionbased}
Dongfang Yang, Ekim Yurtsever, Vishnu Renganathan, Keith~A. Redmill, and Ümit
  Özgüner.
\newblock A vision-based social distancing and critical density detection
  system for covid-19, 2020.

\bibitem{zhang19_wide_area_crowd_count_groun}
Qi Zhang and Antoni~B. Chan.
\newblock Wide-area crowd counting via ground-plane density maps and multi-view
  fusion cnns.
\newblock In {\em 2019 IEEE/CVF Conference on Computer Vision and Pattern
  Recognition (CVPR)}, page nil, 6 2019.

\bibitem{zhang16_singl_image_crowd_count_multi}
Yingying Zhang, Desen Zhou, Siqin Chen, Shenghua Gao, and Yi Ma.
\newblock Single-image crowd counting via multi-column convolutional neural
  network.
\newblock In {\em 2016 IEEE Conference on Computer Vision and Pattern
  Recognition (CVPR)}, 6 2016.

\end{thebibliography}
}

\end{document}